\RequirePackage{fix-cm}
\documentclass[twocolumn]{svjour3}
\smartqed
\usepackage{graphicx}
\usepackage{makecell}
\usepackage{marvosym}
\usepackage{amsmath,amsfonts,amssymb}
\usepackage{color}
\usepackage{xspace}
\usepackage{xcolor}
\usepackage{array}
\usepackage{multirow}
\usepackage{setspace}
\usepackage{tabularx}
\usepackage{caption}
\usepackage{natbib}
\setcitestyle{authoryear,open={(},close={)}}
\usepackage{url}
\usepackage{tabularx}
\usepackage{hhline}
\RequirePackage{snapshot}
\usepackage{nicefrac}
\usepackage{afterpage}
\usepackage{tikz}
\usepackage{amsmath}
\usepackage{mathtools}
\usepackage{color}
\usetikzlibrary{calc}

\begin{document}
	
	\renewcommand{\floatpagefraction}{2}
	\title{Illumination-Invariant Active Camera Relocalization for Fine-Grained Change Detection in the Wild}
	\author{Nan~Li\textsuperscript{1,2} \and Wei~Feng\textsuperscript{1,2}\Letter  \and Qian~Zhang\textsuperscript{1,2}}
	
	\institute{Nan~Li\at
		\email{linan94@tju.edu.cn}\\
		\and 
		Wei~Feng \Letter (corresponding author) \at
		\email{wfeng@tju.edu.cn}\\
		\and
		Qian~Zhang \at
		\email{qianz@tju.edu.cn}\\
		\and
		1\quad The College of Intelligence and Computing, Tianjin University, Tianjin, China.\\
		2\quad The Key Research Center for Surface Monitoring and Analysis of Cultural Relics (SMARC), State Administration of Cultural Heritage, China.
	}
	
	
	\maketitle

	\begin{abstract}
		Active camera relocalization (ACR) is a new problem in computer vision that significantly reduces the false alarm caused by image distortions due to camera pose misalignment in fine-grained change detection (FGCD). 
		Despite the fruitful achievements that ACR can support, it still remains a challenging problem caused by the unstable results of relative pose estimation, especially for outdoor scenes, where the lighting condition is out of control, i.e., the twice observations may have highly varied illuminations. 
		This paper studies an illumination-invariant active camera relocalization method, it improves both in relative pose estimation and scale estimation. 
		We use plane segments as an intermediate representation to facilitate feature matching, thus further boosting pose estimation robustness and reliability under lighting variances.
		Moreover, we construct a linear system to obtain the absolute scale in each ACR iteration by minimizing the image warping error, thus, significantly reduce the time consume of ACR process, it is nearly $1.6$ times faster than the state-of-the-art ACR strategy. 
		Our work greatly expands the feasibility of real-world fine-grained change monitoring tasks for cultural heritages. 
		Extensive experiments tests and real-world applications verify the effectiveness and robustness of the proposed pose estimation method using for ACR tasks.
	\end{abstract}
	
	\section{Introduction}
\begin{figure}
		\centering
		\includegraphics[width=0.475\textwidth]{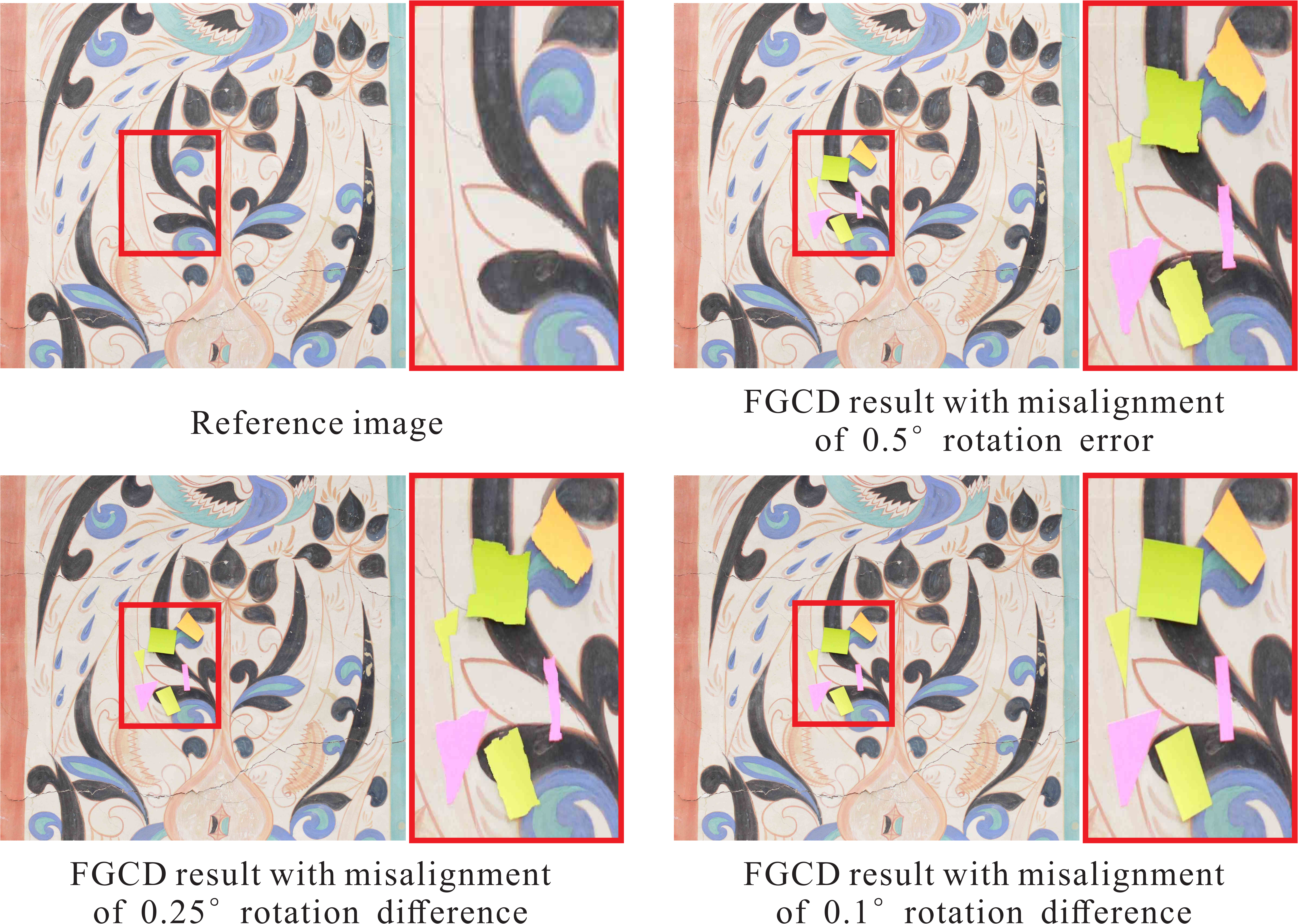}
		\caption{The effect of images' physical alignment on FGCD results. The real man-made fine-grained changes are regular geometries. When there is a deviation greater than $0.1 ^{\circ}$ between the two images, the FGCD algorithm contorts those changes to irregular shapes. Note the camera is $ 50 $ centimeters away from the object, which is the parameter we use in actual work.}
		\label{fig:FGCD_error}
	\end{figure}

	Active camera relocalization (ACR) is a new but classical problem in computer vision, it physically and precisely relocates the current camera to the same pose of the reference one, by actively adjusting the camera pose \citep{ACRPAMI}. 
	This approach that physical align observations is designed to support the Fine-grained change detection (FGCD) task, it aims to detect very fine-grained changes between high-value scenes over long intervals. 
	To guarantee the faithfulness of FGCD results and avoid false alarms caused by distortion from the image alignment algorithm \citep{SCR,SCR1}, we need to obtain highly consistent image observations between two collections via ACR process. 
	Hence the relocalization precision of ACR is very important, which indeed relies on the accuracy of relative camera pose estimation. 
	As shown in Fig~\ref{fig:FGCD_error}, even if the difference between two observations is only $0.5 ^{\circ}$, it still harm the authenticity of detected fine-grained changes.
	
	For real-world ACR problems, both accuracy and robustness of pose estimation are critical, especially for outdoor scenes with large illumination change.
	Despite existing ACR strategies \citep{ACRCVPR,ACRPAMI,ACRmiao} work well in indoor scenes, it cannot work well in outdoor scenes where lighting condition is fully out of control, e.g., cultural heritages or vital equipment in the wild. 
	This is mainly due to the straightforward SIFT or other traditional image feature matching throughout the whole image domain. 
	It results in insufficient and unreliable point correspondences under large illumination change, thus further harming the accuracy of the pose estimation results.
	
	With the rapid development and impressive success of deep learning, researchers recently utilized convolutional neural networks for relative pose estimation problems.
	One approach is to learn to predict the relative pose between images \citep{ummenhofer2017demon,relocnet,laskar2017camera}, the second type is to train models for interest point detectors and descriptors fit for multiple-view feature matching \citep{superpoint,D2-net,yi2016lift}. 
	As evaluated in~\citep{learnlimit}, such data-driven approaches, though, may alleviate the adverse effect of illumination change problems to some extent, yield less accurate results compared to the traditional 3D structure-based methods. 
	Moreover, the generalization capability of learning-based method is highly correlated with the training data or limited on some specific scenes, whose performance, however, is not satisfactory on various real-world scenes. 
	
	Therefore, an urgent and challenging requirement is how to obtain reliable and stable ACR results under the complexity and unrestricted environment in the wild, that is, the lighting condition is fully not of control. To this end, we need to estimate the relative pose as accurately as possible under such conditions. 
	In addition, due to the lack of absolute motion scale in pose estimation, the state-of-the-art ACR methods \citep{ACRCVPR,ACRPAMI} rely on iterative adjustments, such as bisection approaches, by guessing the motion scale and gradually reducing it to approach the target camera pose. Such a guessing process harms the efficiency of ACR process.
	
	In this paper, we propose an illumination-invariant active camera relocalization method (i$^2$ACR) to eliminate the drawbacks discussed above. 
	Firstly, based on a known physical robotic motion in an initialization module, we construct a linear system to obtain the absolute motion scale in each ACR iteration by minimizing the image warping error.
	Secondly, image features like SIFT reflect the local gradient distribution of the image. For the planar region in the image, lighting changes will change the pixel value but has a limited influence on the gradient distribution. 
	Thus, for each image pair requested to estimate relative camera pose, we extract the plane regions from the two images, respectively. 
	After that, we use plane segments as an intermediate representation to facilitate feature matching, thus further boosting pose estimation robustness under lighting changes, while maintaining high accuracy.
	
	Our contributions are as follows:
	1) We present a fast and robust ACR algorithm i$^2$ACR, it performs faster than the state-of-the-art ACR strategy and shows superior performance under scenes with large illumination.
	3) We conduct extensive experiments to verify the feasibility and efficiency of the proposed method. Our approach extends the ACR to more tough real-world scenes, greatly promotes the preventive conservation of cultural heritages in the wild environment.
	
	
	\section{Related Work} \label{sec:relatedwork}
	
	\subsection{Active Camera Relocalization}
	Active camera relocalization, a critical step in the fine-grained change detection task \citep{FGCD}, is proposed to overcome the image distortion caused by the virtual image alignment \citep{ACRPAMI}. Unlike Active recurrence of lighting (ALR), which is used to actively reproduce the lighting conditions of the rephotograph \citep{ALR}, ACR employs a mechanical platform to physically adjust the camera itself to return to the position in which the reference image was taken. 
	So far, ACR has been widely used in the study of history \citep{wells2012western}, monitoring of the natural environment \citep{guggenheim2006inconvenient}, or high-value scenes (e.g., cultural heritage) \citep{FGCD}.
	
	Early works focus on generating navigation information to assist users to adjust camera pose, such as the computational rephotography method \citep{bae2010computational} and the homography-based active camera relocalization method \citep{FGCD}. Through well designed fast matching strategy (including effective keyframe decision and feature substitution), Shi et al. \citep{ACRmobile} proposed a fast and reliable ACR approach that works well on mobile devices. 
	Recently, Tian et al. \citep{ACRCVPR,ACRPAMI} proposed a hand-eye calibration-free ACR strategy from a single 2D reference image, using a commercial robotic platform to automatically adjust the camera pose. Besides, they theoretically prove the convergence of their ACR strategy and achieve $0.1$mm level camera relocalization precision in real-world applications. After that, Miao et al. \citep{ACRmiao} propose a depth camera based ACR strategy to further improve the robustness of ACR under textureless scenes. Despite the diversity and successes of previous ACR strategies, their convergence highly depends on the consistent lighting conditions, without which the foundation of relocalization convergence and accuracy is undermined.

	
	\subsection{Relative Camera Pose Estimation}
	\textbf{Feature-based pose estimation.}
	In the decades, relative camera pose estimation method based local feature, such as SIFT \citep{SIFT}, SURF \citep{SURF}, ORB \citep{ORB}, have been proposed. Typical approaches based on feature matching includes 5-point solver \citep{5pt}, 8-point solver \citep{8pt} etc. Although SIFT claims an advantage that it is robust to lighting changes when it first come out. Over the years of SLAM and SfM applications, SIFT do not performs satisfactorily for pose estimation under the highly variant illuminations.
	
	\textbf{Learning-based pose estimation.}
	For a long time, researchers have been trying to learn the relative camera pose estimation by using neural networks. 
	In the study of visual localization \citep{vislualocalization} and other tasks, some methods that directly regress the relative pose for two input images are proposed \citep{ummenhofer2017demon,relocnet,laskar2017camera} to cooperate with image retrieval \citep{imageretrieval} to achieve absolute pose estimation \citep{posenet}.
	However, due to the complex conversion relationship from image space to $ \mathrm{SE}\left(3 \right) $, the Learning-based method that directly learns the relative pose from images has been difficult to achieve good results. 
	
	\textbf{Hybrid pose estimation.}
	Another way to estimate relative pose is to train models for interest point detectors and descriptors fit for multiple-view feature matching \citep{superpoint,D2-net,yi2016lift}.
	These hybrid methods, which only learns image feature extraction and matching and leave the pose calculation to the epipolar constraint, show excellent performance in challenging image feature matching in specific dataset and provide reliable support for pose estimation in applications such as SLAM and SfM.
	However, the performance of these methods on unknown data will degrade, it do not achieve the same level of pose accuracy as Feature-based methods \citep{zhou2019learn,learnlimit}. 
	Thus, these hybrid methods do not support the relative pose estimation accuracy requirement, e.g., an error less than  $0.1 ^{\circ}$ rotation, in ACR process, which is a high-level standard in applications as SLAM.

	\subsection{Fine-Grained Change Detection}
	Fine-grained change detection~(FGCD) \citep{FGCD}, designed to detect minute changes (pixel-level) between high-value scenes over long intervals. 
	Unlike change detection problems~(CD) that focus on object-level changes in the range of observations \citep{CD1,CD2}, FGCD accomplish such tasks as fine-damage detection of precision instruments, minute-change detection of cultural relics, etc. 
	The changes detected in these applications are often too subtle to be considered noise by CD.
	
	Feng et al. first study the FGCD problem and present a feasible hybrid optimization scheme to separate the true fine-grained change map from the changes caused by varied imaging conditions \citep{FGCD}. 
	After that, Active camera relocalization (ACR) \citep{ACRCVPR,ACRPAMI} makes FGCD simple and reliable by physically restoring the camera's pose. 
	However, in outdoor scenes, ACR is extremely unreliable due to completely uncontrollable lighting conditions, makes FGCD difficult or even fail.


	\section{Illumination-Invariant Active Camera Relocalization (i$^{2}$ACR)}
	
	In this section, we propose a robust and fast ACR method to accurately align the observations under uncontrollable lighting conditions. 
	
	\begin{figure*}
		\centering
		\includegraphics[width=0.85\textwidth]{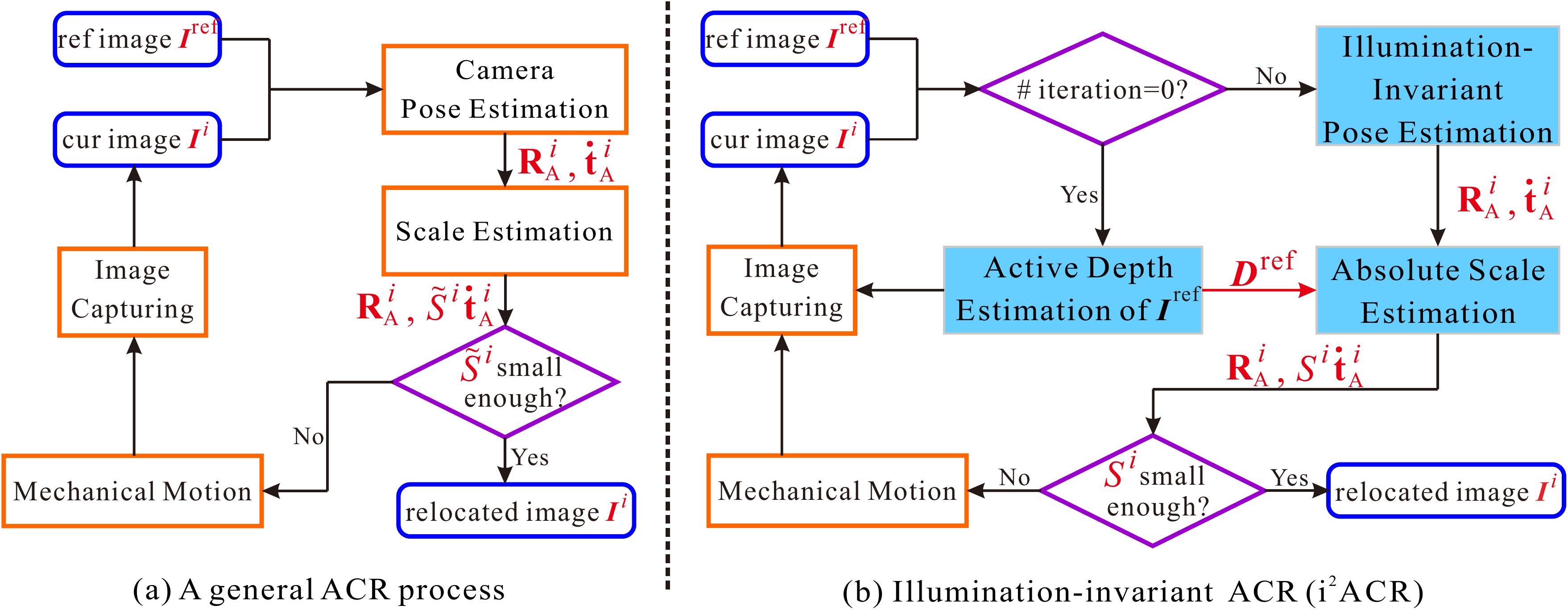}
		\caption{Working flow of the Active Camera Relocalization (ACR) process. (a) and (b) show the framework of a general ACR process and the proposed Illumination-Invariant ACR (i$^2$ACR), respectively.}
		\label{fig:ACR_flowchart}
		\vspace{-0cm}
	\end{figure*}
	
	\subsection{Preliminaries}
	\textbf{Notation.} In this paper, we use a 3D rotation $ \textbf{R}\in \mathrm{SO}\left(3 \right) $ and a 3D translation $ \textbf{t} \in \mathbb{R}^3 $ to indicate the pose $\textbf{P}\in \mathrm{SE}\left(3 \right)$ or the rigid body motion $\textbf{M}\in \mathrm{SE}\left(3 \right)$.
	i.e., $ \textbf{P} = \left[ {\begin{array}{*{20}{c}} \mathbf{R}&\mathbf{t} \\  {\mathbf{0}^\mathrm{T}}&1 \end{array}} \right] \backsimeq \left\langle \textbf{R},\textbf{t} \right\rangle $, the same to $\textbf{M}$. Specifically, $\bf\dot {t}$ denotes the direction of the translation.
	Additionally, considering the inevitable hand-eye relative pose $\textbf{X}\in \mathrm{SE}\left(3 \right)$ ($ \textbf{X} \backsimeq \left\langle \textbf{R}_{\rm X},\textbf{t}_{\rm X} \right\rangle $) in this problem, we use subscripts A and B to indicate the eye and the hand system, respectively.
	Specifically, ${\bf{P}}_{\rm A}\backsimeq \left\langle {\bf{R}}_{\rm A},{\bf{t}}_{\rm A} \right\rangle$ denotes the camera pose, whilst $\bf{P}_{\rm B}\backsimeq \left\langle {\bf{R}}_{\rm B},{\bf{t}}_{\rm B} \right\rangle$ represents the hand pose. Similarly, the rigid body  motions $\bf{M}_{\rm{A}}$ and $\bf{M}_{\rm{B}}$ are named in the same way.
	
	\textbf{A general ACR process.} Essentially, ACR seeks to identify optimal hand motion $ {\bf{M}}_{\rm B}^{*} $ to relocate the camera from its initial pose $ \bf{P}_{\rm A}^{\rm 0} $ to the reference pose $ \bf{P}_{\rm A}^{\rm ref} $,
	\begin{equation}\label{ACR formu}
		\bf{M}_{\rm B}^{\rm *}  = \mathop{\arg\min}_{\textbf{M}\in \mathrm{SE}\left(\rm 3 \right)} \| \bf{X}\bf{M}\bf{X}^{\rm -1}\bf{P}_{\rm A}^{\rm 0}-\bf{P}_{\rm A}^{\rm ref}\|_{\rm F}^{\rm 2},
	\end{equation}
	where $ \| \cdot \|_{\rm F} $ is the Frobenius norm.
	It seems that all that is required is a reliable relative camera pose. However, it is expensive to frequently obtain the relative hand-eye pose in practice.
	Thus, the researchers~\citep{ACRPAMI} propose a general ACR process, as shown in Fig~\ref{fig:ACR_flowchart}. They guess the $ \bf X $ by $ \tilde{\bf X} = \bf I $ and execute a series of hand motions to achieve the equivalent effect of executing $ \bf{M}_{\rm B}^{\rm *} $,
	\begin{equation}\label{ACR formu in step}
		{\bf{M}_{\rm B}^{ *}}=\prod_{i=n}^{0}{\tilde{\bf{M}}_{\rm B}}^{ n}{\tilde{\bf{M}}_{\rm B}}^{n-1}\cdots{\tilde{\bf{M}}_{\rm B}}^{i}\cdots{\tilde{\bf{M}}_{\rm B}}^{0},
	\end{equation}
	where ${\tilde{\bf{M}}_{\rm B}}^{i} $ is the physically real state of the $i$-th guessed hand motion. 
	We know $({\bf{P}}_{\rm A}^{i})^{-1}\tilde{\bf{X}} = \tilde{\bf{X}}\tilde{\bf{M}}_{\rm B}^{i}$, where ${{\bf{P}}_{\rm A}^{i}}\backsimeq \left\langle {\bf{R}}_{\rm A}^i,{\bf{t}}_{\rm A}^i \right\rangle $ and $\tilde{\bf{M}}_{\rm B}^{i}\backsimeq \left\langle \tilde{\bf{R}}_{\rm B}^{i},\tilde{\bf{t}}_{\rm B}^{i} \right\rangle $. Thus, when $\tilde{\bf X} = \textbf{I}$, ${\tilde{\bf{M}}_{\rm B}}^{i} $ can be expressed as
	\begin{equation}\label{ACR formu in Rt}
		{\tilde{\bf{R}}_{\rm B}}^{i} = ({\bf{R}}_{\rm A}^{i})^{-1}, \quad  \quad {\tilde{\bf{t}}_{\rm B}}^{i} = -({\bf{R}}_{\rm A}^{i})^{-1}{\bf{t}_{\rm A}}^{ i}.
	\end{equation}
	The authors of~\citep{ACRPAMI} prove that if the ACR process Eq.~(\ref{ACR formu in step}) iterates enough times, i.e., $n$ becomes large enough, it will satisfactorily converge.
	
	
	\subsection{Overview}
	Compared with the general ACR process, our i$^2$ACR process (see Fig~\ref{fig:ACR_flowchart}) has improved in terms of both estimation and camera pose estimation. 
	Based on a known physical robotic motion in an initialization module, we construct a linear system to obtain the absolute motion scale in each ACR iteration by minimizing the image warping error (\textit{c.f.} Sec.~\ref{sec:Initialization for Absolute Scale}). 
	To estimate relative camera pose for each image pair requested, we extract the plane regions from the two images, respectively. 
	We then use plane segments as an intermediate representation to facilitate feature matching, thus further boosting the pose estimation robustness under lighting changes, whilst maintaining a high degree of accuracy (\textit{c.f.} Sec.~\ref{sec:ii PE}).


	\subsection{Initialization for Absolute Scale Estimation}\label{sec:Initialization for Absolute Scale}
\begin{figure}
		\centering
		\includegraphics[width=0.425\textwidth]{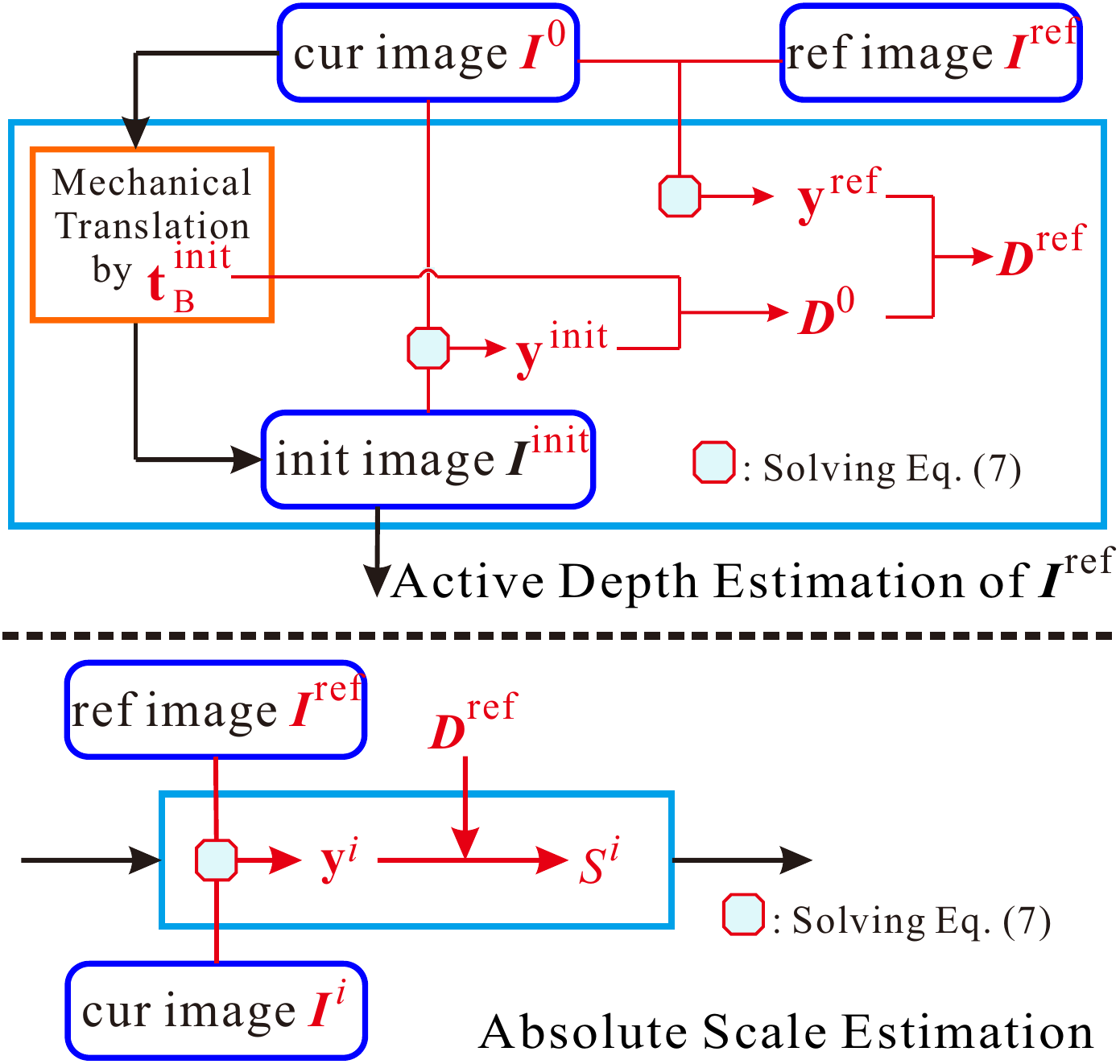}
		\caption{Illustration of the Active Estimation of Depth and Absolute Scale Estimation modules of our i$^2$ACR method (see Fig.~\ref{fig:ACR_flowchart}).}
		\label{fig:initialization}
	\end{figure}

	By using a reference image ${\textbf{\textit{I}}}^{\rm ref}$ taken in the past and a current image ${\textbf{\textit{I}}}^{ i}$ taken at this moment only, we can obtain the missing-scale relative camera pose, i.e., ${\bf\dot{P}}_{\rm A}^{i}\backsimeq \left\langle {\bf{R}}_{\rm A}^{i},{\bf\dot{t}_{\rm A}}^{ i} \right\rangle $. 
	As set out in Fig.~\ref{fig:initialization}, we introduce an initialization module to acquire the absolute scale ${{S}^{i}}$ for ${\bf\dot{P}}_{\rm A}^{i}$ in each ACR iteration. 
	In the following, we elaborate on the key steps for estimating ${{S}^{i}}$.
	
	\subsubsection{Ratio of absolute scale to depth}
	For simplicity, we assume the first camera pose is the world coordinate system. Subsequently, the relationship between the homogeneous coordinate point ${\bf q}^{a} \in \mathbb{R}^3$ on ${\textbf{\textit{I}}}^{a}$ and the corresponding 3D point ${\bf Q}^{a} \in \mathbb{R}^3$ in world coordinate can be expressed as
	\begin{equation}\label{qQ1}
		{\bf Q}^{a} = {\bf K}^{-1}{\bf q}^{a}D^{a},
	\end{equation}
	where ${\bf K}$ is the camera intrinsic matrix and $D^{a}$ is the depth of ${\bf q}^{a}$. By taking ${\bf{P}_{\rm A}}\backsimeq \left\langle {\bf{R}}_{\rm A},{\bf\dot{t}_{\rm A}}S \right\rangle $ as the camera relative pose between ${\textbf{\textit{I}}}^{a}$ and a second image ${\textbf{\textit{I}}}^{b}$, we then get the 3D point ${\bf Q}^{b} \in \mathbb{R}^3$ in world coordinate by
	\begin{equation}\label{qQ2}
		{\bf Q}^{b} = {{\bf{R}}_{\rm A}}^{-1}{\bf K}^{-1}{\bf q}^{b}D^{b} - {{\bf{R}}_{\rm A}}^{-1}{\bf\dot{t}_{\rm A}}S,
	\end{equation}
	where ${\bf q}^{ b} \in \mathbb{R}^3 $ is the homogeneous coordinate point on ${\textbf{\textit{I}}}^{b}$ and $D^{b}$ indicates the depth of ${\bf q}^{b}$. The matching homogeneous coordinate points between ${\textbf{\textit{I}}}^{a}$ and ${\textbf{\textit{I}}}^{b}$ correspond to the same 3D point ${\bf Q}$. Hence, minimizing the image warping error between ${\textbf{\textit{I}}}^{a}$ and ${\textbf{\textit{I}}}^{b}$ produces
	\begin{equation}\label{argmin}
		\{ \hat D^{a}_{ i}, \hat D^{b}_{ i}, \hat{S} \} = \mathop{\arg\min}_{ D^{a}_{ i},  D^{b}_{ i}, {S} } \dfrac{1}{2} \sum_{i=1}^N \| {\bf Q}^{a}_{ i} - {\bf Q}^{b}_{ i}\|_{\rm F}^{\rm 2},
	\end{equation}
	where $ \| \cdot \|_{\rm F} $ is the Frobenius norm and $N$ is the number of matching feature points between ${\textbf{\textit{I}}}^{a}$ and ${\textbf{\textit{I}}}^{b}$. By integrating Eqs.~(\ref{qQ1})--(\ref{argmin}), the relationship between depth and absolute scale can be expressed as
	\begin{equation}\label{ratio}
		\{ \hat D^{a}_{ i}, \hat D^{b}_{ i}, \hat{S} \} = \mathop{\arg\min}_{ D^{a}_{ i},  D^{b}_{ i}, {S} } \dfrac{1}{2} \sum_{i=1}^N \mathrm{F} \, (D^{a}_{ i},  D^{b}_{ i},{S}),
	\end{equation}
	where the detailed expression of function $\mathrm{F}(\cdot)$ is derived in the Appendix. Minimizing $\mathrm{F}(\cdot)$ w.r.t. $D^{a}_{ i}$, $D^{b}_{ i}$ and ${S}$ resulting in a $3N \times (2N + 1)$ homogeneous linear system, i.e., Eq.~(\ref{linear system}) in the Appendix, whose solution is
	\begin{equation}\label{minimum non-zera solution}
		{\bf y}=\left[ d^{a}_{ 1},d^{b}_{ 1},..., d^{a}_{ i},d^{b}_{ i},...,d^{a}_{ N},d^{b}_{ N} ,{s}\right],
	\end{equation}
	Note, $\bf y$ is the minimum non-zero solution of Eq.~(\ref{linear system}) and the lowercase $d$ and $s$ are not the depth and absolute scale values. The ratio between $y_{2N+1}$ and $y_i, (i=1,\ldots,2N)$ represents the ratio between the absolute scale $ S$ and depth $ D$, where $y_i $ is the $j$-th element in ${\bf y}$.
	
	\subsubsection{Active depth estimation of ${\textbf{\textit{I}}}^{\, \rm ref}$ }
	At the very beginning of an ACR process, as shown in Fig~\refeq{fig:initialization}, we first execute a known rigid body translation ${\bf{M}}_{\rm B}^{\rm init} \backsimeq \left\langle {\bf{I}},{{\bf{t}}_{\rm B}^{\rm init}} \right\rangle $ and capture an intermediate image ${\textbf{\textit{I}}}^{\rm init}$. 
	Then we estimate the relative camera pose ${{\bf\dot{P}}_{\rm A}^{\rm init}}\backsimeq \left\langle {\bf{I}},{{\bf\dot{t}}_{\rm A}^{\rm init}} \right\rangle $ of ${\textbf{\textit{I}}}^{\rm init}$ and the current image ${\textbf{\textit{I}}}^{0}$. 
	Proceeding on this basis, the absolute scale $S^{\rm init}$ of ${\bf\dot{P}}_{\rm A}^{\rm init}$ can be calculated by $S^{\rm init} = \|{{\bf{t}}_{\rm B}^{\rm init}}\|/\|{\bf\dot{t}}_{\rm A}^{\rm init}\|$. 
	In section~\ref{sec:analysis X}, we show that the hand-eye relative pose $\textbf{X}$ \textit{does not affect} the calculation of $S^{\rm init}$, meaning that it \textit{does not affect} the calculation of $S^{i}$ in the subsequent ACR iterations.
	
	Based on Eqs.~(\ref{argmin})--(\ref{ratio}), we construct a linear system from ${\textbf{\textit{I}}}^{0}$ and ${\textbf{\textit{I}}}^{\rm init}$ and then derive the solution
	\begin{equation}
		{\bf y}^{\rm init}\!=\!\left[  d^{0}_{ 1},d^{\rm init}_{ 1},...,d^{0}_{ j},d^{\rm init}_{j},..., d^{0}_{J},  d^{\rm init}_{ J}, {s}^{\rm init}\right],
	\end{equation}
	where $J$ is the number of the matching feature points between ${\textbf{\textit{I}}}^{0}$ and ${\textbf{\textit{I}}}^{\rm init}$. As is apparent, we can get the sparse depth map ${\textit{\textbf D}}^{0}$ of the current image  ${\textbf{\textit{I}}}^{0}$ through
	\begin{equation}\label{curdepth}
		\hat D^{0}_{ j} = d^{0}_{ j}~\frac{{S}^{\rm init}}{{s}^{\rm init}} .
	\end{equation}
	Similarly, for the reference image ${\textbf{\textit{I}}}^{\rm ref}$ and the current image ${\textbf{\textit{I}}}^{0}$, we can obtain the relation
	\begin{equation}\label{final ratio}
		{\bf y}^{\rm ref}=\left[  d^{\rm ref}_{ 1},d^{0}_{ 1},...,d^{\rm ref}_{l},d^{0}_{l} ,..., d^{\rm ref}_{ L},d^{0}_{L},  {s}^{0}\right],
	\end{equation}
	where $L$ is the number of the matching feature points between ${\textbf{\textit{I}}}^{\rm ref}$ and ${\textbf{\textit{I}}}^{0}$, $L \leq J$. 
	Hence, with the sparse depth map ${\textit{\textbf D}}^{0}$ and Eq.~(\ref{final ratio}), we finally obtain a reference sparse depth map ${\textit{\textbf D}}^{\rm ref}$ through
	\begin{equation}\label{refdepth}
		\hat D^{\rm ref}_{ l} = \hat D^{0}_{ l}~\frac{d^{\rm ref}_{ l}}{{d^{0}_{ l}}} .
	\end{equation}
	This depth map of ${\textbf{\textit{I}}}^{\rm ref}$ is the basis for the calculation of absolute scale in the subsequent ACR process.

	
	\subsubsection{Absolute scale estimation in each ACR iteration}
	The state-of-the-art ACR algorithm 5pt-ACR \citep{ACRPAMI} guesses a scale $ \tilde{S}^i $ for $ {\bf\dot{t}}^{i}_{\rm A} $ through a bisection approaching strategy. Guessing a scale $\tilde{S}^i $ for the $i$-th hand motion is indeed a reasonable solution, however, it usually takes a number of iterations to achieve ACR converge. As a result, a lot of time is spent on mechanical motion and image computation in practical applications.
	With our initialization module, the number of ACR iterations can be greatly reduced by calculating rather than guessing the absolute scale.
	
	For the reference image ${\textbf{\textit{I}}}^{\rm ref}$ and a current image ${\textbf{\textit{I}}}^{i}$ in the $i$-th ACR iteration ($i \geq 1$), we have the relation
	\begin{equation}\label{ACR iteration ratio}
		{\bf y}^{i}\!=\! \left[  d^{\rm ref}_{ 1},d^{i}_{ 1},...,d^{\rm ref}_{p},d^{i}_{p} ,..., d^{\rm ref}_{ P},d^{i}_{P},  {s}^{i}\right],{\rm{s.t.}}\; P \! \leq \! L,
	\end{equation}
	where $P$ is the number of the matching feature points between ${\textbf{\textit{I}}}^{i}$ and ${\textbf{\textit{I}}}^{\rm ref}$. 
	With the sparse depth map ${\textit{\textbf D}}^{\rm ref}$, we can calculate the absolute scale ${S}^i$ by
	\begin{equation}\label{absolute_scale_in_ACR}
		S^i = \frac{1}{P}( \sum_{p=1}^P  \frac{s^i  \hat D^{\rm ref}_{ p}}{d^{\rm ref}_{ p}}) .
	\end{equation}
	Thus, we derive the relative camera pose ${\bf{P}}_{\rm A}^{i}\backsimeq \left\langle {\bf{R}}_{\rm A}^{i},S^i {\bf\dot{t}_{\rm A}}^{ i} \right\rangle $ and then execute the $i$-th hand motion $\tilde{\bf{M}}_{\rm B}^{i}$ according Eq.~(\ref{ACR formu in Rt}). Following multiple adjustments, ACR can be considered finished if $S^i$ is small enough.
	
	Although our strategy still requires multiple iterations to converge due to the unknown hand-eye relative pose $\bf X$, providing the exact absolute scale of the camera relative pose can expedite convergence considerably. Compared with 5pt-ACR, our approach can effectively reduce the iteration number of the ACR process from more than ten times to less than four times without compromising accuracy. See the experimental section for a detailed comparison and discussion of this point.
	
	\subsubsection{Analysis of the influence of hand-eye calibration}\label{sec:analysis X}
	In this part, we will demonstrate that in the initialization module, even if we do not know the hand-eye relative pose $\textbf{X}$, it will not affect the accuracy of $S^{\rm inint}$ calculated by $S^{\rm init} = \|{{\bf{t}}_{\rm B}^{\rm init}}\|/\|{\bf\dot{t}}_{\rm A}^{\rm init}\|$.
	For convenience, we assume ${\bf{P}}_{\rm A}^{\rm init}$ is the relative pose between the current eye (camera) pose and initial eye (camera) pose while ${\bf{P}}_{\rm B}^{\rm init}$ is the relative pose between the current hand pose and initial hand pose.
	
	By referring to the hand-eye relative pose $\textbf{X}$, we can obtain the relations $ {\bf{X}}{\bf{M}}_{\rm B}^{\rm init} = {\bf{M}}_{\rm A}^{\rm init} {\bf{X}} = ({\bf{P}}_{\rm A}^{\rm init})^{-1} {\bf{X}}$~\citep{ACRPAMI}. Hence, we have
	\begin{equation}\label{scale X}
		\begin{aligned}
			\hat {\bf{M}}_{\rm B}^{\rm init}  &= \mathop{\arg\min}_{\textbf{M}\in \mathrm{SE}\left(\rm 3 \right)} \| {\bf{X}}{\bf{M}}{\bf{X}}^{-1}{\bf{P}}_{\rm A}^{\rm init}-{\bf{I}}\|_{\rm F}^{\rm 2} \\
			&= {\bf{X}}^{-1} ({\bf{P}}_{\rm A}^{\rm init})^{-1} {\bf{X}},
		\end{aligned}
	\end{equation}
	where $ \| \cdot \|_{\rm F} $ is the Frobenius norm and $\bf{I} $ is the unit matrix, ${{\bf{P}}_{\rm A}^{\rm init}}\backsimeq \left\langle {\bf{I}},{{\bf{t}}_{\rm A}^{\rm init}} \right\rangle $. We split Eq.~(\ref{scale X}) into its rotation and translation parts, which yields
	\begin{equation}\label{scale X R}
		{\bf{I}} = ({\textbf{R}}_{\rm X})^{-1}{\bf{I}}{\textbf{R}}_{\rm X},
	\end{equation}
	\begin{equation}\label{scale X T}
		\begin{aligned}
			{\bf{t}}_{\rm B}^{\rm init} &= ({\textbf{R}}_{\rm X})^{-1}{\bf{I}}{\bf{t}}_{\rm X} - ({\textbf{R}}_{\rm X})^{-1}{\bf{I}}{{\bf{t}}_{\rm A}^{\rm init}} - ({\textbf{R}}_{\rm X})^{-1}{\bf{t}}_{\rm X} \\
			&= - ({\textbf{R}}_{\rm X})^{-1}{{\bf{t}}_{\rm A}^{\rm init}} .
		\end{aligned}
	\end{equation}
	According to Eq.~(\ref{scale X T}), it can reasonably be concluded that absolute scale $S^{\rm init}$ can be calculated by $\|{{\bf{t}}_{\rm B}^{\rm init}}\| = \|{{\bf{t}}_{\rm A}^{\rm init}}\| = S^{\rm init} \|{{\bf\dot{t}}_{\rm A}^{\rm init}}\|$ without first knowing the hand-eye relative pose. More specifically, by executing a known rigid body translation, we can accurately estimate the absolute scale $S^{i}$ in the subsequent ACR iterations without knowing the hand-eye relative pose.
	
	
	\subsection{Illumination-Invariant Pose Estimation}\label{sec:ii PE}
\begin{figure}
		\centering
		\includegraphics[width=0.45\textwidth]{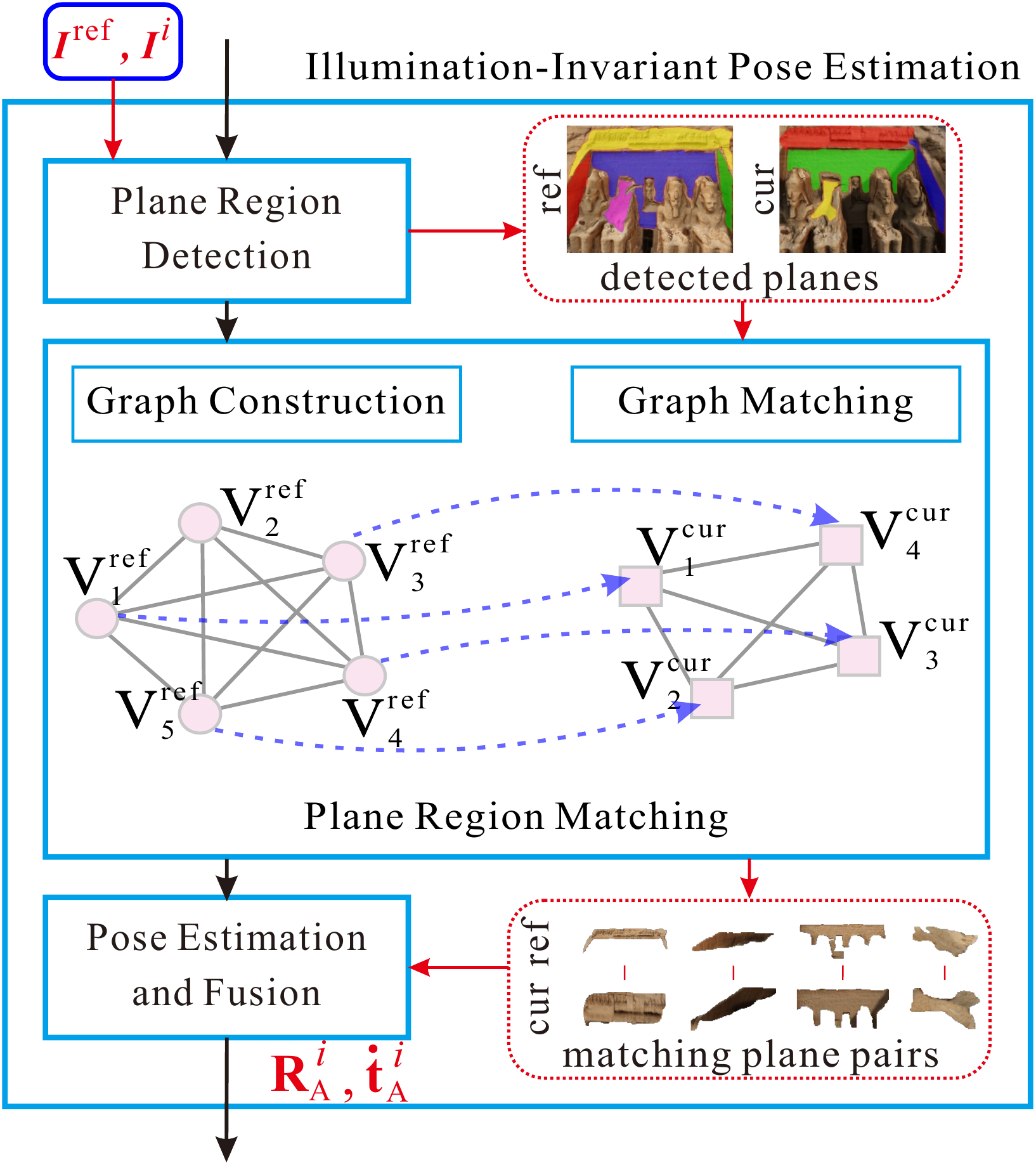}
		\caption{Illustration of the Illumination-Invariant Pose Estimation (i$^2$PE) module in Fig.~\ref{fig:ACR_flowchart}.}
		\label{fig:IIPE}
	\end{figure}

	For the two images ${\textbf{\textit{I}}}^{\rm ref}$ and ${\textbf{\textit{I}}}^{\rm cur}$ taken under different lighting conditions, as shown in Fig.~\ref{fig:IIPE}, we use planes as intermediate representations to effectively improve the robustness and reliability of pose estimation. As has been shown, image features such as SIFT reflect the local gradient distribution of the image. For the planar region in the image, lighting changes will change the pixel value, although they will have a limited influence on the gradient distribution. Due to this, the planar structure and the matching between plane regions can eliminate the adverse effects of large illumination variances (see experimental results for the verification). In this subsection, we introduce the proposed illumination-invariant pose estimation (i$^2$PE) and describe in detail how to use planes as intermediate representations to boost pose estimation robustness under different illuminations.
	
	\subsubsection{Plane region detection}
	We apply Plane R-CNN~\citep{planercnn} to extract the plane regions from ${\textbf{\textit{I}}}^{\rm ref}$ and ${\textbf{\textit{I}}}^{\rm cur}$, respectively. Plane R-CNN employs a variant of Mask R-CNN~\citep{mask} to detect planes and performs best in the plane detection task. However, there are still inevitable errors near the plane boundary of the detection results. Since the feature descriptors near the plane boundaries are subjected to lighting changes. In such conditions, improving the segmentation accuracy on plane boundaries does not make much sense. Therefore, we subjected the detected planes to mild morphological erosion~\citep{erosion} to further improve pose estimation accuracy.
	

	\subsubsection{Plane region matching}
	Let ${{\mathcal{I}}^{\rm ref}} = \{ {\textbf{\textit P}}_h^{{\mathop{\rm ref}\nolimits} }\}_{h=1}^{H}$ be the collection of $ H $ planes detected from ${\textbf{\textit{I}}}^{\rm ref}$, with ${\textbf{\textit P}}_h^{{\mathop{\rm ref}\nolimits} } $ being the $h$-th detected plane. Similarly, let ${{\mathcal{I}}^{\rm cur}} = \{ {\textbf{\textit P}}_m^{{\mathop{\rm cur}\nolimits} }\}_{m=1}^{M}$ be the collection of $ M $ planes detected in ${\textbf{\textit{I}}}^{\rm cur}$.
	To find the matching relations implied in ${\mathcal{I}}^{\rm ref}$ and ${\mathcal{I}}^{\rm cur}$, i.e., the same plane region captured by the two images, we model it as a graph matching problem. As shown in Fig.~\ref{fig:IIPE}, for ${\mathcal{I}}^{\rm ref}$, the node-set $ {\mathcal{V}}^{\rm ref} $ of the undirected graph ${\mathcal{G}}^{\rm ref}=\left\lbrace{\mathcal{V}}^{\rm ref}, {\mathcal{E}}^{\rm ref} \right\rbrace $ denotes the $ H $ detected planes. The undirected edge-set $ {\mathcal{E}}^{\rm ref} $ indicates the minimum Euclidean distance between any two detected planes in $ \mathcal{I}^{\rm ref} $. We construct undirected graph ${\mathcal{G}}^{\rm cur}$ from ${\mathcal{I}}^{\rm cur}$ in the same way.
	Then, matching the two graphs ${\mathcal{G}}^{\rm ref}$ and ${\mathcal{G}}^{\rm cur}$ is to identify an optimal node-to-node correspondence ${\bf{ U}}\in \left\lbrace1,0 \right\rbrace^{H\times M} $, where ${\bf{U}}_{ac} = 1$ when the nodes $V_a^{\rm ref} \in {\mathcal{V}}^{\rm ref}$ and $V_c^{\rm cur} \in {\mathcal{V}}^{\rm cur}$ are matched, ${\bf{U}}_{ac} = 0$ for otherwise. In this paper, the assignment matrix $ \bf{U} $ has the following condition,
	\begin{equation}
		\label{condition}
		\mathcal{U}\!\!=\!\!\left\lbrace\!{\bf{U}}\! \mid \!{\bf{U}}\!\in\! \left\lbrace1,0 \right\rbrace^{H\times M},{\bf{U}}^{\top} \! \textbf{1}_H \!\leq\!\textbf{1}_M, {\bf{U}}\textbf{1}_M \!=\!\textbf{1}_H \!\right\rbrace\!,
	\end{equation}
	where ${\bf{1}}_M$ and ${\bf{1}}_H$ are unit vectors with dimension $M$ and $H$, respectively. Eq.~(\ref{condition}) constrains the plane matching problem to be a one-to-one mapping model. Considering ${\mathcal{G}}^{\rm ref}$ and ${\mathcal{G}}^{\rm cur}$ may have the different sizes, we use the inequality in Eq.~(\ref{condition}) to indicate the assumption that $ H \leq M $.
	
	The matching problem can be solved by maximizing an objective function that measures the node and edge affinities between ${\mathcal{G}}^{\rm ref}$ and ${\mathcal{G}}^{\rm cur}$. As such, we use the potential $w_v(V_a^{\rm ref} , V_c^{\rm cur})  $ to measure the similarity between nodes $V_a^{\rm ref}$ and $V_c^{\rm cur}$. Additionally, we use the potential $w_e(E_{ab}^{\rm ref} , E_{cd}^{\rm cur}) $ to measure the similarity between edges $E_{ab}^{\rm ref}$ and $E_{cd}^{\rm cur}$.
	Specifically, $w_v(V_a^{\rm ref} , V_c^{\rm cur})  $ is measured by the number of matching feature points between $ {\textbf{\textit P}}_a^{\mathop{\rm ref}\nolimits}$ and $ {\textbf{\textit P}}_c^{\mathop{\rm cur}\nolimits}$, $w_e(E_{ab}^{\rm ref} , E_{cd}^{\rm cur}) $ is measured by the difference between the minimum Euclidean distance of $ {\textbf{\textit P}}_a^{\mathop{\rm ref}\nolimits}$ to $ {\textbf{\textit P}}_b^{\mathop{\rm ref}\nolimits}$ and $ {\textbf{\textit P}}_c^{\mathop{\rm cur}\nolimits}$ to $ {\textbf{\textit P}}_d^{\mathop{\rm cur}\nolimits}$.
	
	We integrate these two types of similarities through an affinity matrix ${\bf{W}}\in{\mathbb{R}}^{MH\times MH}$, in which the diagonal element corresponds to the unary potential $w_v(V_a^{\rm ref} , V_c^{\rm cur})  $ and the non-diagonal element corresponds to the pairwise potential $w_e(E_{ab}^{\rm ref} , E_{cd}^{\rm cur}) $. Based on the description above, we generate the objective function
	\begin{equation}\label{gmformu}
		\begin{aligned}
			{\bf{U}_{\rm c}^{\top}\bf{W}\bf{U}_{\rm c}}\!=\!{\sum_{\mathclap{U_{ac}=1}} \! w_v(V_a^{\rm ref} , V_c^{\rm cur})}\!+\!{\sum_{\mathclap{\substack{U_{ac}=1\\U_{bd}=1}}} \! w_e(E_{ab}^{\rm ref} , E_{cd}^{\rm cur} )},
		\end{aligned}
	\end{equation}
	where $\bf{U}_{\rm c}$ is the column expansion of $\bf{U}$. Then, the plane region matching problem can be addressed by identifying the optimal assignment matrix ${\bf{U}}^{*}$ 
	\begin{equation}
		\label{opt}
		{\bf{U}}^{*}=\mathop{\arg\max}_{{\bf{U}}} {\bf{U}}_{\rm c}^{\top}{\bf{W}}{\bf{U}}_{\rm c}, \ \ \rm{s.t.}\; {\bf{U}} \in \mathcal{U}.
	\end{equation}
	We use the deformable graph matching method~\citep{soveGM} to solve Eq.~(\ref{opt}) and then obtain the $K$ matching plane region pairs $ {\mathcal{M}} = \{(  {\textbf{\textit P}}_i^{\mathop{\rm ref}\nolimits}, {\textbf{\textit P}}_i^{\mathop{\rm cur}\nolimits} ) \}_{i=1}^{K}$ from the optimal assignment matrix ${\bf{U}}^{*}$.
	
	\begin{table*}[t]
		\caption{Time performance and AFD values for $8$ scenes. The red values represent failed relocalizations. See text for details.}
		\vspace{-0.2cm}
		\label{tab:ACR_quantitive_results}
		\centering
		\fontsize{7pt}{12pt}\selectfont
		\renewcommand\tabcolsep{8.55pt}
		\begin{spacing}{1.25}
			\begin{center}
				\begin{tabular*}{0.9475\textwidth}[c]{ l |  c c c | c  c c }
					\Xhline{1pt}
					\multirow{2}{*}{{Scene}}
					&\multicolumn{3}{c|}{5pt-ACR} 	&\multicolumn{3}{c}{i$^{2}$ACR}	  \\ \cline{2-7}
					&AFD	 &Iteration times		&Time (s)	  &AFD	 &Iteration times		&Time (s) \\  	\hline				
					constantL1 & {\bf0.809} & 13.4 & 254.6 & 0.821 & {\bf3.4} & {\bf100.6} 	\\
					constantL2 & 1.077 & 13.8 & 266.8 & {\bf1.002} & {\bf3.2} & {\bf104.8} \\
					constantL3 & 0.519 & 13.2 & 259.6 & {\bf0.486} & {\bf3.0} & {\bf105.2} \\
					constantL4 & {\bf0.765} & 13.6 & 261.2 & 0.793 & {\bf2.8} & {\bf101.8} \\ \hline	
					constantL\_average& 0.793 & 13.5 & 260.6 & 0.776 & 3.1 & 103.1 \\ \hline	
					variedL1 & \color{red}143.513 & 16.6 & 298.4 & {\bf0.972} & {\bf2.8} & {\bf97.8} \\
					variedL2 & \color{red}76.749 & 17.4 & 321.6 & {\bf1.003} & {\bf3.8} & {\bf103.8} \\
					variedL3 & \color{red}161.333 & 16.2 & 289.2 & {\bf0.690} & {\bf3.0} & {\bf100.6}\\
					variedL4 & \color{red}96.401 & 17.8 & 323.4 & {\bf0.851} & {\bf3.2} & {\bf104.2} \\ \hline	
					variedL\_average& \color{red}119.499 & 17.0 & 308.2 & 0.879 & 3.2 & 101.6 \\
					\Xhline{1pt}							
				\end{tabular*}
			\end{center}
		\end{spacing}\vspace{-0.2cm}
	\end{table*}

	\subsubsection{Pose estimation and fusion}
	Given a matching plane pair $ (  {\textbf{\textit P}}_i^{\mathop{\rm ref}\nolimits}, {\textbf{\textit P}}_i^{\mathop{\rm cur}\nolimits} ) $, the camera relative pose can be estimated by decomposing the homography matrix calculated from ${\textbf{\textit P}}_i^{\mathop{\rm ref}\nolimits}$ and ${\textbf{\textit P}}_i^{\mathop{\rm cur}\nolimits}$. Compared to the other feature-based pose estimation algorithms, e.g., 5-point solver~\citep{5pt}, which is essentially unaffected by the planar degeneracy and works, the decomposing homography matrix~\citep{faugeras1988motion} offers obvious advantages in relation to noise interference (see experimental results for the verification).
	
	In theory, the $ K $ relative poses estimated from the $K$ matching plane pairs in $ \mathcal{M} $ are the same. We fuse those that exhibit variance according to their reliability to further ensure accuracy and robustness. In this regard, reliability can be considered from two aspects:~$1)$~\emph{Number}, we prefer the estimated pose from the plane pair which has numerous matching feature points.~$2)$~\emph{Distribution}, we prefer the estimated pose from the plane pair which has a more uniform distribution of feature points.
	
	By using planes as an intermediate representation, we obtain reliable and accurate pose estimation results under highly variant illuminations. In addition, the matching between the planes does not consider those variation areas, such that, our method can also effectively eliminate the interference of the partial region changes on camera pose estimation.


	\section{Experimental Results}
	
	\subsection{Setup}
	In the ACR verification experiments, we compare the proposed illumination-invariant active camera relocalization (i$^{2}$ACR) with the state-of-the-art ACR strategy ($5$pt-ACR) \citep{ACRPAMI}. 
	We build $8$ different scenes to execute the ACR process on the monitoring platform with a Canon $5$D MarkIII camera. 
	These scenes can be divided into $4$ scenes with varied lighting conditions (variedL$ 1 $--$ 4 $), and $4$ normal scenes with constant lighting conditions (constantL$ 1 $--$ 4 $). 
	We use the average feature-point displacement (AFD)~\cite{ACRPAMI} to quantitatively measure the relocalization accuracy,
	\begin{equation}\label{AFD}
		{\rm AFD}({\textbf{f}}_{\rm ref}-{\textbf{f}}_{\rm cur})=\frac{1}{l}\sum_{i=1}^l \| {\textbf{f}}_{\rm ref}^{\,i}-{\textbf{f}_{\rm cur}^{\,i} \|_{\rm 2}},
	\end{equation}
	where $ {\textbf{f}}_{\rm ref} $ and $ {\textbf{f}}_{\rm cur} $ are the matching feature-point coordinates in the reference image and the current relocalized image, respectively. Meanwhile, $ l $ is the number of matches. 
	In the following, we will set out a detailed explanation and analysis of each experiment.

	
	\subsection{Comparison of i$^{2}$ACR with the state-of-the-art}
\begin{figure*}
		\centering
		\includegraphics[width=0.695\textwidth]{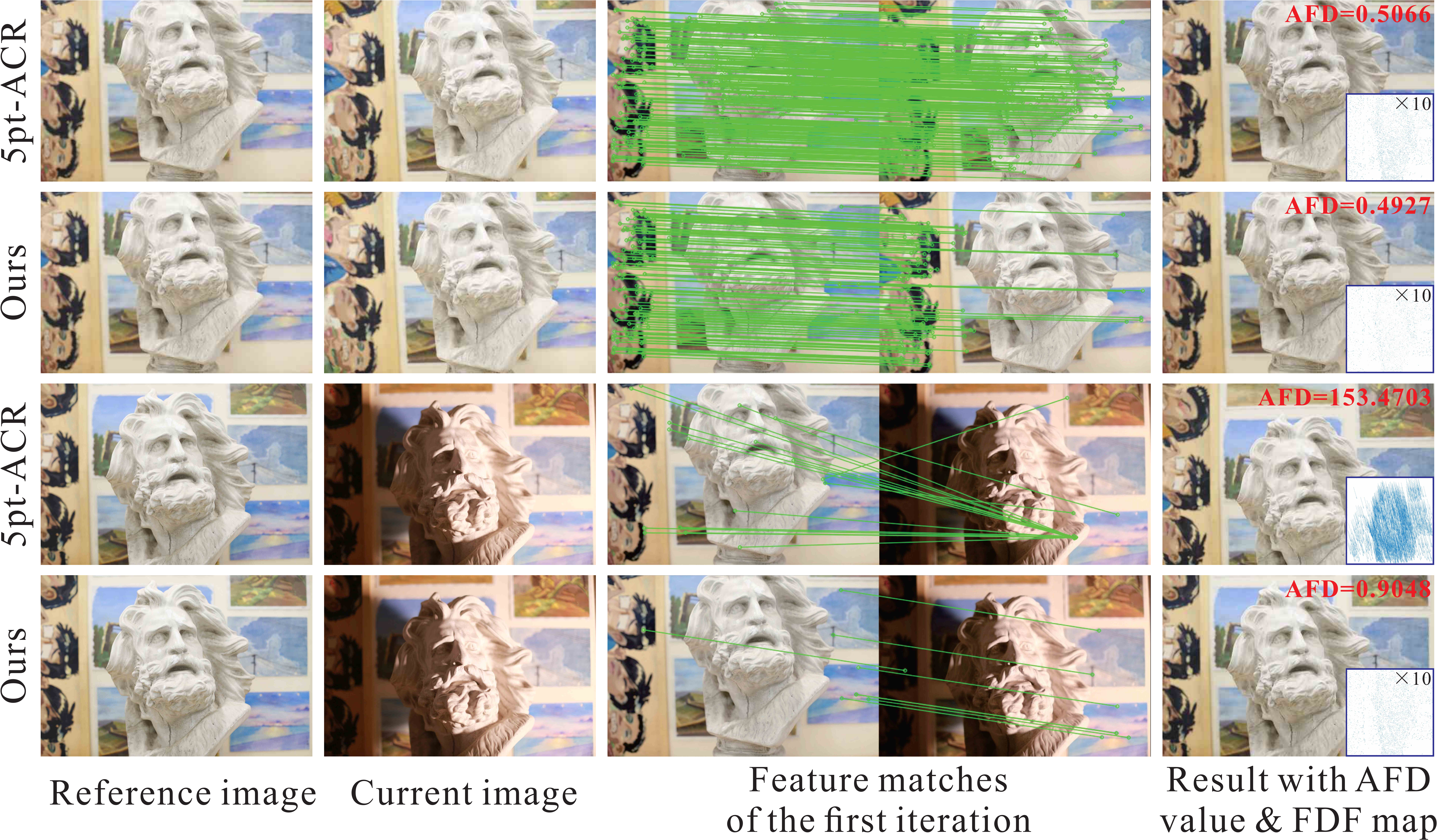}
		\caption{Two groups of ACR with (the third and fourth rows) and without (the first and second rows) lighting difference, respectively. In the third column, we visualize the feature matches at the first iteration of each ACR process. We restore the lighting conditions to the same as the reference when ACR is finished, then we calculate the AFD values of each ACR process, and the results are shown in the fourth column.}
		\label{fig:ACRexample}
	\end{figure*}

	For $5$pt-ACR and our i$^{2}$ACR algorithm, we perform ACR for each algorithm five times for each of the $8$ scenes, i.e., constantL$ 1 $--$ 4 $ and variedL$ 1 $--$ 4 $. 
	To guarantee fairness, we also reproduce the reference and initial camera poses for each independent test of the same scene. 
	Fig.~\ref{fig:ACRexample} gives the visual results of the ACR two groups in constantL$1$ (the first two rows) and variedL$1$ (the last two rows). 
	Detailed time performance and average AFD values are listed in Table~\ref{tab:ACR_quantitive_results}.

	\subsubsection{Robustness analysis}
	Table~\ref{tab:ACR_quantitive_results} shows that the accuracy of our method is comparable to the $5$pt-ACR's in constantL$ 1 $--$ 4 $, which is reasonable. 
	As shown in the first two rows of Fig.~\ref{fig:ACRexample}, compared to $5$pt-ACR, i$^{2}$ACR only reduces the number of feature matches involved in pose estimation. Moreover, there are almost no mismatched feature matches after RANSAC~\citep{RANSAC} when the lighting conditions do not change. From this, it can be concluded that the pose estimation accuracy of both $5$pt-ACR and ours can satisfactorily support the ACR process. 
	
	Compared to the $5$pt-ACR strategy used in variedL$ 1 $--$ 4 $, ours demonstrated greater superiority. 
	By using the $5$pt-ACR strategy, the ACR process failed in all experiments under variedL$ 1 $--$ 4 $. 
	As shown in the third row of Fig.~\ref{fig:ACRexample}, sharp differences in lighting change the appearance of scenes (especially scenes featuring $ 3 $D structure), hence the feature point descriptors involved in pose estimation change, which give rise to the dramatic mismatching of feature points even after RANSAC. 
	Thus, the accuracy performance of $5$pt-ACR degrades significantly when the two observations have highly varied illuminations. 
	In contrast, i$^{2}$ACR shows the same accuracy performance in variedL$ 1 $--$ 4 $ as in constantL$ 1 $--$ 4 $. 
	The fourth row of Fig.~\ref{fig:ACRexample} visualizes an example of feature matching in the first iteration of i$^{2}$ACR executed in variedL$ 1$.
	As can be seen from Fig.~\ref{fig:ACRexample} and Table~\ref{tab:ACR_quantitive_results}, since planes are used as an intermediate representation to qualify the search space for feature matching, i$^{2}$ACR exhibits a significant advantage in accuracy and robustness, thus, effectively supporting the valid ACR process.
	
	
	\subsubsection{Convergence rate analysis}
\begin{figure}
		\centering
		\includegraphics[width=0.475\textwidth]{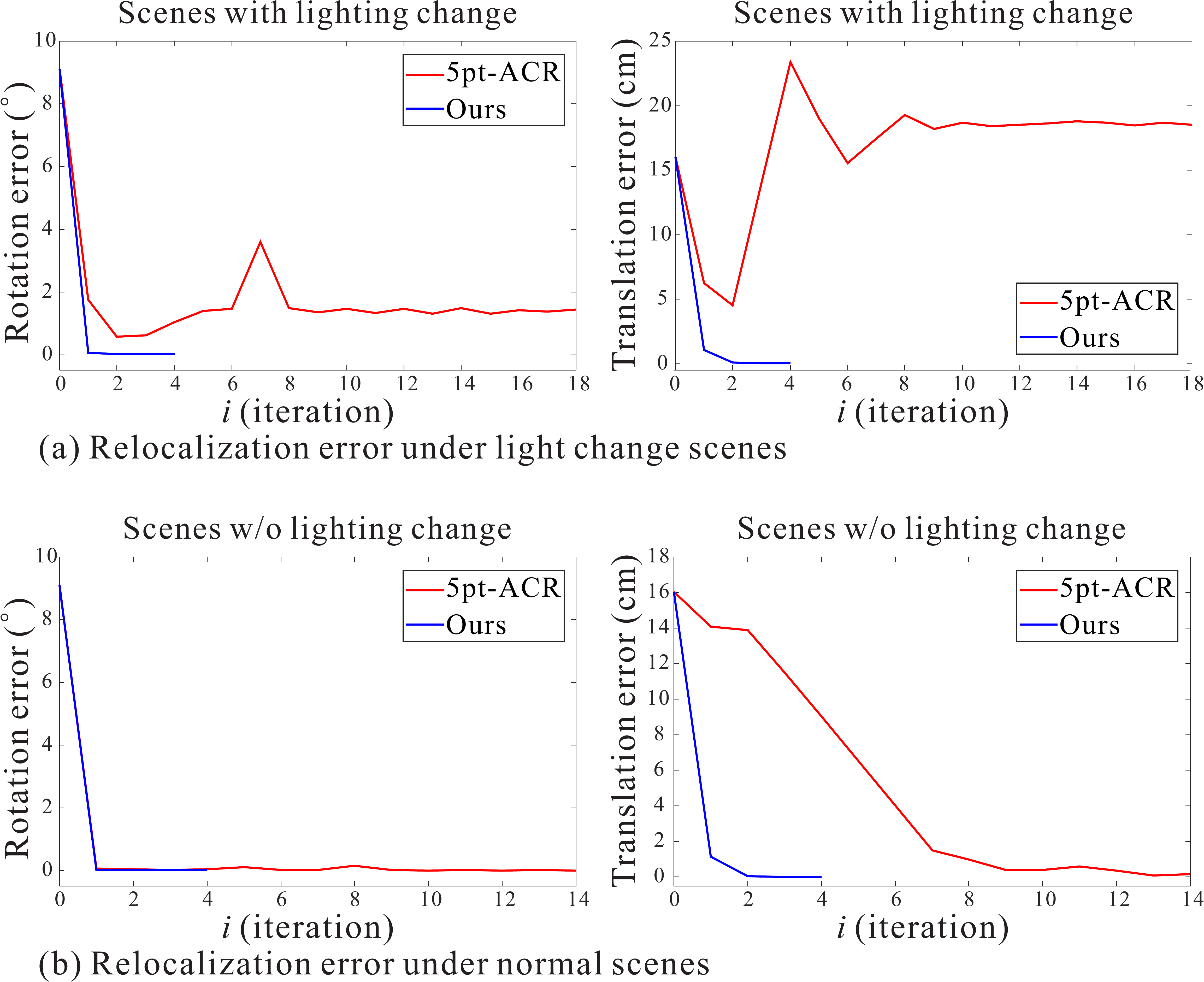}
		\caption{Average convergence process of $ {\bf{R}}_{\rm A} $ and $ {\bf{t}}_{\rm A} $ in constantL$ 1 $--$ 4 $ and variedL$ 1 $--$ 4 $. (a) shows the convergence curves under scenes with lighting change while (b) shows the convergence curves under scenes without lighting change.}
		\label{fig:virRTerror}
	\end{figure}

	We analyze all of the ACR experiments and average the results in constantL$ 1 $--$ 4 $ and variedL$ 1 $--$ 4 $, respectively. 
	As shown in Fig.~\ref{fig:virRTerror}, the variation trend of relocalization error (rotation difference and translation error) shows the ACR convergence process. Specifically, we use Euler angles in three axes to denote the rotation of relative pose for convenience.
	When considering only the performance of effective ACR processes (experiments under constantL$ 1 $--$ 4 $), Fig.~\ref{fig:virRTerror}(b) demonstrates that i$^{2}$ACR converges $ 3.5 $ times faster than the state-of-the-art $5$pt-ACR, Table~\ref{tab:ACR_quantitive_results} shows that the real application time will be reduced by $ 60\% $ compared with the state-of-the-art $5$pt-ACR.
	This reduction is largely due to our method provides the absolute scale for each ${\bf{M}}_{\rm B}^{i}$, thus expediting the convergence of Eq.~(\ref{ACR formu in step}). By way of contrast, $5$pt-ACR relies on guessing the translation scale, which requires more adjustments to converge.
	
	Note, because of the unknown relative hand-eye $\textbf{X}$, which we treat it as identity matrix $\textbf{I}$, and the imperfect pose estimation, ACR cannot guarantee camera pose convergence in just one step. As such, it takes several iterations to eliminate the influence of $\textbf{X}$ and imperfect pose estimation results and ensure that ACR converges, see \citep{ACRPAMI} for details. 
	However, as can be seen from Table~\ref{tab:ACR_quantitive_results}, the iteration number (average $3.15$ times) of i$^{2}$ACR is much less than the iteration number of $5$pt-ACR (average $15.25$ times).
	
	\begin{figure*}
		\centering
		\includegraphics[width=0.95\textwidth]{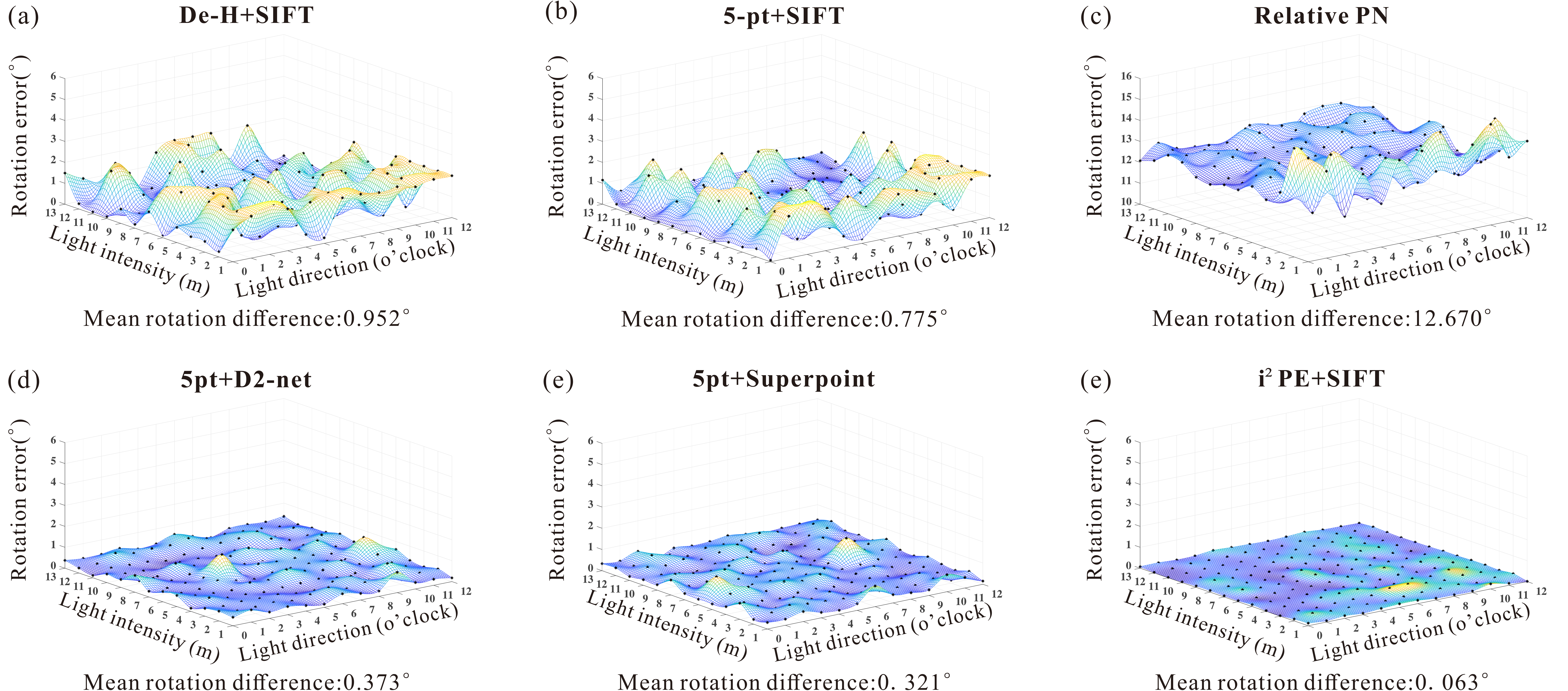}
		\caption{(a)--(f) shows the representative comparison results of our pose estimation method and baselines under varied lighting intensities and directions, the smoother the heat-map, the more robust the method. Note that our method performs superior to baselines for accuracy and robustness.}
		\label{fig:virtualexp}
	\end{figure*}

	\subsection{Analysis of Relative Pose Estimation}
	As noted above, the ACR relocalization precision relies on the accuracy of relative camera pose estimation. 
	In this subsection, we further explore and verify the effectiveness of i$^{2}$ACR by comparing the relative pose estimation performance of the proposed pose estimation method in i$^{2}$ACR with that of other pose estimation method in situations with varying lighting.
	
	
	\subsubsection{Datasets and baselines}
	\textbf{Datasets.}
	To evaluate the proposed i$^{2}$PE on challenging data exhibiting a large illumination difference, we use Unity 3D to create a virtual environment and quantitatively control the changes in lighting intensity and direction. 
	Specifically, we design $ 13 $ varied lighting directions, i.e., $ 1,...,12 $ o'clock side lightings and front lighting. 
	We also design $ 13 $ different lighting intensities; the change in lighting intensity is equivalently achieved by controlling the distance between the light source and the scene. 
	For convenience, we represent lighting intensity in terms of distance in this paper. 
	In particular, we treat the front lighting direction with the intensity of $7$ meters as the standard environment.

	In this lighting-controlled virtual environment, we construct a dataset consisting of $ 510 $ images of three building models ($170 \times 3 $). 
	For each building, we capture a reference image in the standard environment and then control the virtual camera to execute a certain motion $\bf{M}_{\rm A}$. 
	By adjusting the lighting direction ($ 1,...,12 $ o'clock directions and the front direction) and intensity ($1,2,...,13$ m), we take $169$ current images under each intensity and each direction condition. 
	Ultimately, we generate $507$ reference-current image pairs from the dataset.
	
	\textbf{Metrics and baselines.}
	In this paper, we employ the angular error ($^{\circ}$) to measure the correction of the estimated rotation. 
	For the estimated translation, we use the angular error ($^{\circ}$) to measure the correction of its direction. 
	Our baselines for comparison are as follows, ranging from purely hand-crafted to purely data-driven models.
	\textbf{Feature-based}: (1)\textit{De-H} \citep{faugeras1988motion}\textit{ + SIFT}, (2)\textit{5-point solver} \citep{5pt}\textit{ + SIFT}. 
	\textbf{Learning-based}: (3)\textit{Relative PN} \citep{laskar2017camera}. 
	\textbf{Hybrid}: (4)\textit{De-H + D2-net} \citep{D2-net}, (5)\textit{5-point solver + D2-net}, (6)\textit{De-H + Superpoint} \citep{superpoint}, (7)\textit{5-point solver + Superpoint}. 
	For convenience, we call 5-point solver 5-pt in the rest of this paper.
	
	\begin{table*}[t]
		\caption{Comparisons of i$^2$PE and baselines. We also test the performance of using Superpoint and D2-net instead of SIFT in i$^2$PE, respectively}
		\vspace{-0.2cm}
		\label{tab:absolute scale}
		\centering
		\fontsize{7pt}{12pt}\selectfont
		\renewcommand\tabcolsep{5.55pt}
		\begin{spacing}{1.25}
			\begin{center}
				\begin{tabular*}{0.785\textwidth}[c]{ c  l|  c  c   | c  c  }
					\Xhline{1pt}
					\multicolumn{2}{c|}{\multirow{2}{*}{method}}
					&\multicolumn{2}{c|}{w/o lighting difference} &\multicolumn{2}{c}{with lighting difference}	  \\ \cline{3-6}
					& &$\textbf{R}$ error($^{\circ}$)	 &$\bf\dot{t}$ error($^{\circ}$)			  &$\textbf{R}$ error($^{\circ}$)	 &$\bf\dot{t}$ error($^{\circ}$)  \\  			\hline
					\multirow{2}{*}{Feature-based} 
					&De-H+SIFT  &0.269  &2.377  &0.952  &4.006 \\
					&5-pt+SIFT  &\textbf{0.004}  &\textbf{0.082}  &0.775  &5.532 \\
					\hline
					\multirow{4}{*}{Hybrid} 
					&De-H+Superpoint  &0.920  &0.778  &1.138  &0.721 \\
					&5-pt+Superpoint  &0.308  &0.108  &0.321  &0.155 \\
					&De-H+D2-net  &1.208  &1.991  &2.032  &1.837 \\
					&5-pt+D2-net  &0.300  &0.801  &0.373  &2.159 \\
					\hline
					Learning-based
					&Relative PN  &12.371  &11.744  &12.670  &11.623 \\
					\hline
					\multirow{3}{*}{Ours} 
					&i$^{2}$PE+Superpoint  &0.311  &0.108  &0.323  &0.155 \\
					&i$^{2}$PE+D2-net  &0.351  &0.922  &0.357  &1.966 \\
					&i$^{2}$PE+SIFT  &\textbf{0.004}  &0.089  &\textbf{0.063}  &\textbf{0.097} \\
					\Xhline{1pt}							
				\end{tabular*}
			\end{center}
		\end{spacing}\vspace{-0.2cm}
	\end{table*}
	
	Among them, De-H estimates the relative pose by decomposing the homography matrix. 
	In our experiments, we use the implementation of it in ORB-SLAM~\citep{slam1}, a little bit different is that we use SIFT instead of ORB for feature extraction and description. 
	Relative PN is a learning-based approach for camera relocalization. 
	When given a query image, it first retrieves similar database images and then predicts the relative pose between the query and the database images. 
	In our experiments, we treat the reference image as the database image whilst the current image is the query. 
	We specify the reference image (database) corresponding to the current image (query) rather than retrieving it. 
	D2-Net and Superpoint are learning-based method that learns the description and detection of local features. 
	In our experiments, we apply them to replace the SIFT-based feature extraction and description used in De-H+SIFT and 5-pt+SIFT for pose estimation, respectively.
	
	\subsubsection{Less learn is more}
	\textbf{Comparison results.}
	Table~\ref{tab:absolute scale} compares the accuracy of our approaches to the baselines. 
	As can be seen, when there are lighting differences between the reference and current images, i$^{2}$PE+SIFT performs superior to the others. When there is no lighting difference, our method performs almost as well as 5-pt+SIFT, while performing best.
	Fig.~\ref{fig:virtualexp} compares the robustness of our approaches to the baselines by using the rotation error of the poses estimated under varied lighting conditions.
	The result shows that our proposed pose estimation method i$^{2}$PE+SIFT performs best.
	The mean rotation difference is only $ 0.063 ^{\circ} $ and the change in rotation error is almost irrelevant to the illumination variation.
	Referring to Table~\ref{tab:absolute scale} and Fig.~\ref{fig:virtualexp}, with regard to relative camera pose estimation, the Learning-based method and the hybrid method are not particularly sensitive to the illumination changes. Moreover, when there are illumination differences between the reference and current images, the hybrid method even performs better than the Feature-based method. 
	Unfortunately, although this precision may satisfy application requirements such as visual localization, it is far from sufficient for FGCD tasks.
	
	\begin{figure*}
		\centering
		\includegraphics[width=0.995\textwidth]{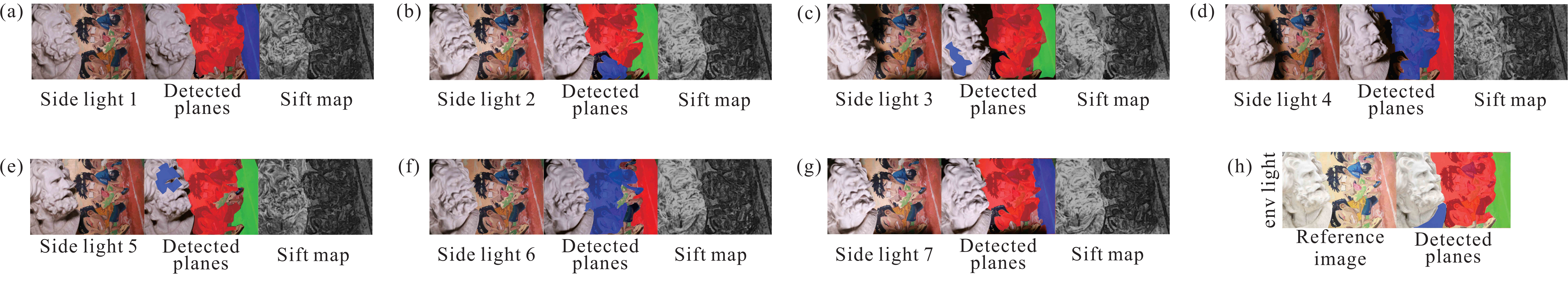}
		\caption{(a)--(g) show the visual results of plane detection and Euclidean distance maps of SIFT feature descriptors between environment lighting image (h) and varied side lightings. Note that the red, blue and green area are the plane detection results.}
		\label{fig:siftmap}
	\end{figure*}
	
	\textbf{What should neural networks learn?} For a long time, researchers have been trying to learn the relative camera pose estimation by using neural networks. 
	However, due to the complex conversion relationship from image space to $ \mathrm{SE}\left(3 \right) $, it is difficult to generate satisfactory results using the Learning-based method that directly learns the relative pose from images. 
	Although the hybrid method, which only learns image feature extraction and matching and leaves the calculation to the epipolar constraint, yields better results than the Learning-based method, it do not achieve the same level of pose accuracy as Feature-based methods \citep{zhou2019learn,learnlimit}.
	
	The overarching advantage of deep learning lies in image understanding tasks such as classification and segmentation.  
	As has been shown, image features like SIFT reflect the local gradient distribution of the image.
	For the planar region in the image, lighting changes will change the pixel value, although they have a limited influence on the gradient distribution. 
	As shown in Fig.~\ref{fig:siftmap}, SIFT feature descriptors located in the plane region are less impacted by lighting changes. Specifically, the Euclidean distance of SIFT feature descriptors which we called sift map in Fig.~\ref{fig:siftmap} denotes the difference between SIFT feature descriptors at the same location in the reference and current images, wherein the brighter the pixel, the bigger the corresponding descriptor difference.
	Thus, we use neural networks to segment the planar region to help pose estimation. 
	In addition, as an intermediate representation in feature matching, effective plane matching further limits the retrieval range in the process of feature-point matching, thus greatly reducing the occurrence of feature-point mismatching.
	
	Strictly speaking, our method should also be viewed as a hybrid method. 
	One of the main reasons why it can achieve better results than D2-Net+5-pt is that the data-driven method experiences generalization problems, as shown in Fig.~\ref{fig:siftmap}. Moreover, Plane R-CNN has difficulties guaranteeing the accuracy of the segmentation results on the plane boundary. Therefore, in the subsequent feature extraction and matching, we abandon the boundary area of the plane segments, thus avoiding the disadvantages flowing from network model generalization to a considerable extent.

	\begin{figure}
		\centering
		\includegraphics[width=0.495\textwidth]{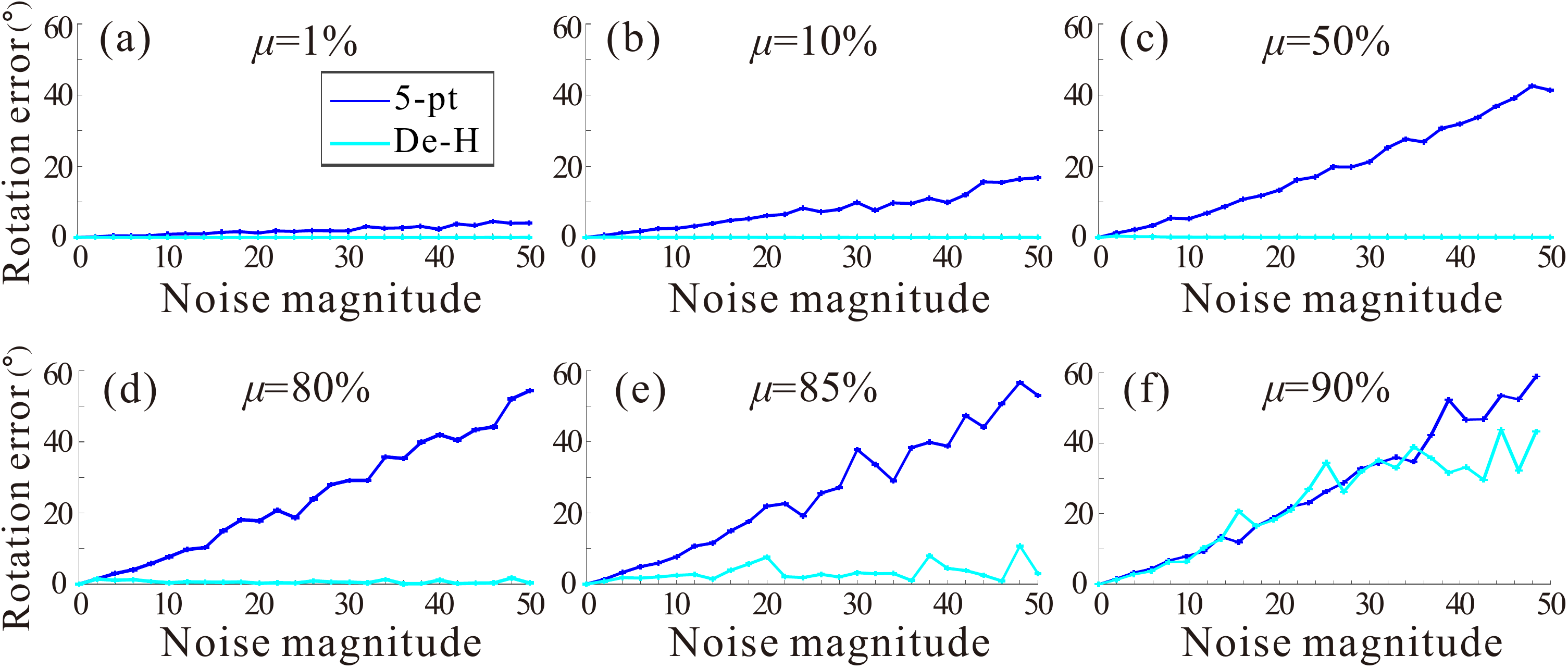}
		\caption{Relation of noise magnitude on matching points and the angle (axis-angle) error, under varied noise ratios.}
		\label{simulation}
	\end{figure}
	
	\subsubsection{Why not use the $5$-point solver}
	$5$pt-ACR uses the \textit{SIFT + 5-point solver} for pose estimation and performs well when applied to pure planar scenes, such as ancient murals. Our i$^{2}$ACR selects \textit{SIFT + De-H} for pose estimation and also performs well. We then design a simulation experiment to verify the camera pose estimated by decomposing the homography matrix from a single plane that is more stable than that estimated by the 5-point solver.

	The experimental data are $1000$ virtual space points in a plane. To simulate the real scene and the real camera as best we could, we set up a virtual camera with the same parameters as Canon $5$D MarkIII. We project space points onto the virtual camera imaging plane and obtain the corresponding reference $2$-D coordinates. Next, we control the virtual camera to execute a rigid body motion and obtain the current $2$-D coordinates of those space points. After that, we add noises ($\mu \times 1000$, $0\leq \mu \leq 100\%$) of different magnitude $\rm r$ which uniformly increases from $0$ to $50$ with a step size of $2$. Finally, we estimate the motion of the virtual camera by 5-pt and De-H, respectively. Note that the noises added in this experiment come from a uniform distribution $ \rm U(-r,r) $.
	
	Fig.~\ref{simulation} shows $6$ results of this experiment. We represent the ground truth and the estimated rotation in terms of the axial angle representation. Since the calculated translation only indicates the direction and does not contain the absolute scale, we only compare the calculated rotation with the ground truth.
	The experimental results show that even though the noise magnitude is slight (e.g., $ 1\%$), the angle estimated by 5-pt is still worse. Meanwhile, when $\mu < 80\%$, De-H is strongly robust to feature matching noises. As for $\mu \geq 90\%$, both the angle estimated by De-H and 5-pt are unreliable. Note that the severe matching errors that invalidate the two methods are reasonable. Hence, in i$^{2}$ACR, we use SIFT+De-H for pose estimation.
	
	
	\begin{figure*}
		\centering
		\includegraphics[width=0.905\textwidth]{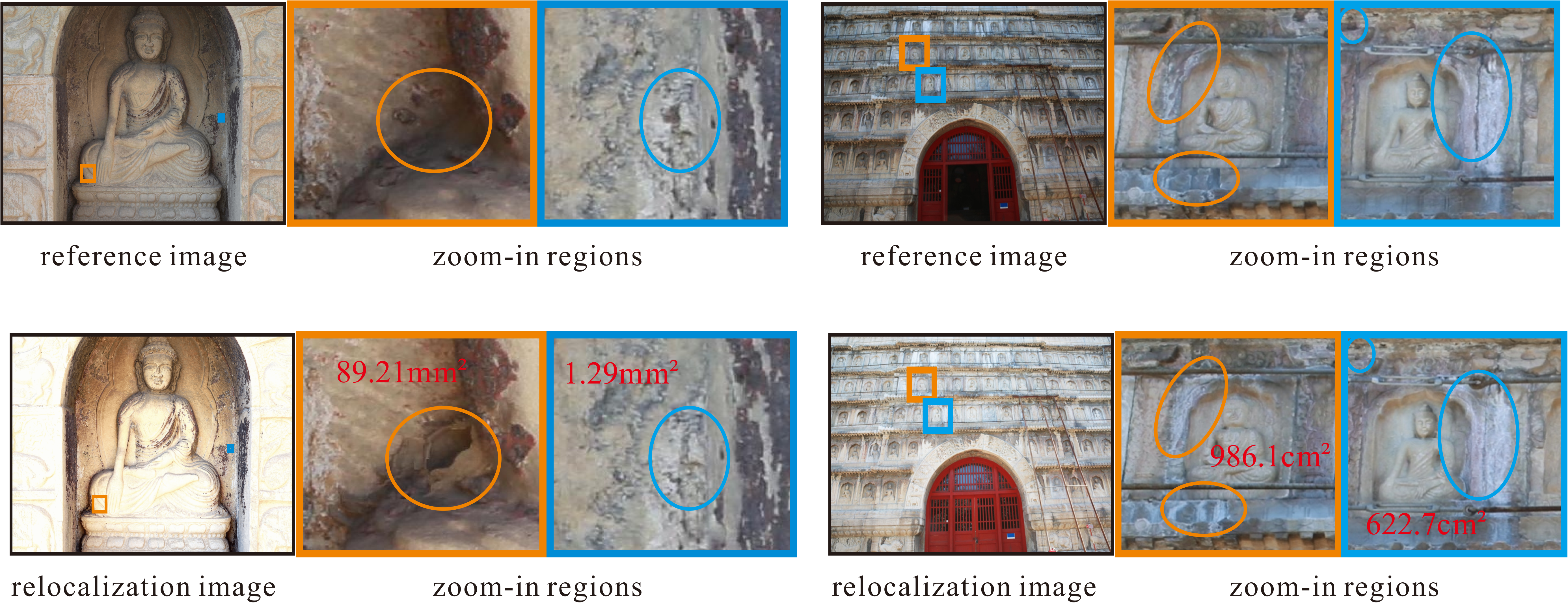}
		\caption{Real-world fine-grained change monitoring cases in the Five-Pagoda Temple. The reference image is captured in June 2019, the relocalized image is captured in August 2019.}
		\label{realworld}
	\end{figure*}

	\subsection{Real-world applications}
	As aforementioned, ACR study is originally motivated by long-time-interval fine-grained change monitoring and measurement of cultural heritages in wild hosting environments. This is a very important and challenging problem that is an essential part for preventive conservation of cultural heritages \citep{ACRPAMI}. This is a very important and challenging problem that is an essential part for preventive conservation of cultural heritages \citep{historical}.
	
	In the last three years, using the proposed i$^{2}$ACR approach, we have conducted monthly fine-grained status inspection of several world cultural heritages in China, including the Palace Museum (PM), the Summer Palace (SP), Dunhuang Mogao Grottoes, Humble Administrator's Garden, and the Five-Pagoda Temple (FPT), covering indoor ancient murals and outdoor ancient buildings, earthen sites stone-sculpted relics.
	
	Table~\ref{tab:realword} shows the average AFD scores for each piece of historical heritage by using $5$pt-ACR and our i$^{2}$ACR strategy, respectively. There are $8$ scenes for each historical heritage. The results in Table~\ref{tab:realword} further illustrate the effectiveness and superiority of i$^{2}$ACR. Note that the monitoring task of the Palace Museum is carried out in a controlled indoor lighting condition, and the outdoor lighting environment of the Summer Palace is essentially the same as when the reference images were taken. Fig.~\ref{realworld} gives some examples of real fine-grained deterioration changes of the Five-Pagoda Temple, discovered by our i$^{2}$ACR strategy from June to August 2019. We also try to use $5$pt-ACR to relocate the camera pose to the reference but failed due to the varied illumination difference compared with the reference. Intuitively, we adjusted the zoom-in regions of the reference and relocalization image in Fig.~\ref{realworld} to make their respective illuminations consistent.
	
	\begin{table}[t]
		\caption{Time performance and AFD error in real-world applications. Note that the PM, the SP, the FPT are the Palace Museum, the Summer Palace, and the Five-Pagoda Temple, respectively.}
		\vspace{-0.2cm}
		\label{tab:realword}
		\centering
		\fontsize{3pt}{12pt}\selectfont
		\begin{spacing}{1.0}
			\begin{center}
				\begin{tabular*}{0.435\textwidth}[c]{ l |  c  c | c   c }
					\Xhline{1pt}
					\multirow{2}{*}{{Scene}}
					&\multicolumn{2}{c|}{5pt-ACR} 	&\multicolumn{2}{c}{i$^{2}$ACR}	  \\ \cline{2-5}
					&AFD	&Time (s)	  &AFD  	&Time (s) \\  	\hline					
					PM & {\bf0.901} &  271.250 & 0.973 &  {\bf106.375} \\
					SP & 2.710 & 262.125  & {\bf1.963} &  {\bf102.750} \\
					FPT &  \color{red}failed & \color{red} failed & {\bf2.005} & {\bf105.125} \\
					\Xhline{1pt}							
				\end{tabular*}
			\end{center}
		\end{spacing}\vspace{-0.2cm}
	\end{table}
	
	Table~\ref{tab:realword} and Fig.~\ref{realworld} fully demonstrate that our proposed method makes up for the lack of outdoor fine-grained change monitoring tasks. This is important as many high-value targets require us to grasp their periodic fine-grained changes.

	
	\section{Conclusion}
	In this paper, we have addressed illumination robustness and convergence speed, two critical issues of active camera relocalization (ACR), by proposing i$^2$ACR, a new fast and robust ACR scheme. Specifically, we first design an initialization module, thus, construct a linear system to obtain the absolute motion scale in each ACR iteration by minimizing the image warping error. Second, we use plane segments as an intermediate representation to facilitate feature matching, thus further boosting pose estimation robustness and reliability under lighting variances. Since the improvement both in relative pose estimation and scale estimation, our i$^2$ACR achieves $1.6\times$ faster convergence speed, much better relocalization accuracy and robustness over state-of-the-art ACR competitors, especially in the wild.
	
	Furthermore, we discuss how to balance the deep learning and the traditional methods in a relative pose estimation pipeline under the premise of pursuing high precision results. As verified by our experiment, image features like SIFT on the planar structures are less subjected to the lighting change. Thus, we use neural networks to segment the planar region to help pose estimation. Compared with the feature-based, learning-based, and hybrid pose estimation methods, ours shows great superiority in both accuracy and generalization.
	
	
	\appendix
	\section{Detailed derivation}
	In this appendix, we show the detailed steps to get the ratio between the depth and the absolute scale from the Eq.~({\ref{qQ1}})$-$(\ref{ratio}). First, we specify the following mathematical representations:
	\begin{equation}\label{A}
		\begin{aligned}
			&\alpha = {{\bf q}^{a}}^{\top} ({{\bf K}^{-1}})^{\top} {{\bf K}^{-1}} {{\bf q}^{a}},\\
			&\beta = {{\bf q}^{a}}^{\top} ({{\bf K}^{-1}})^{\top} {{\bf R}^{-1}} {{\bf K}^{-1}} {{\bf q}^{b}},\\
			&\gamma = {{\bf q}^{a}}^{\top} ({{\bf K}^{-1}})^{\top} {{\bf R}^{-1}} {{\bf\dot{t}_{\rm A}}},\\
			&\delta = {{\bf q}^{b}}^{\top} ({{\bf K}^{-1}})^{\top} ({{\bf R}^{-1}})^{\top} {{\bf R}^{-1}} {{\bf K}^{-1}} {{\bf q}^{b}},\\
			&\epsilon = {{\bf q}^{b}}^{\top} ({{\bf K}^{-1}})^{\top} ({{\bf R}^{-1}})^{\top} {{\bf R}^{-1}} {{\bf\dot{t}_{\rm A}}},\\
			&\zeta = ({{{\bf\dot{t}_{\rm A}}}})^{\top} ({{\bf R}^{-1}})^{\top} {{\bf R}^{-1}} {{\bf\dot{t}_{\rm A}}}.
		\end{aligned}
	\end{equation}
	For the item in Eq.~(\ref{argmin}), we have the equivalent expression:
	\begin{equation}\label{argmin1}
		\| {\bf Q}^{a}_{ i} - {\bf Q}^{b}_{ i}\|_{\rm 2}^{\rm 2} = ({\bf Q}^{a}_{ i} - {\bf Q}^{b}_{ i})\!^{\top}({\bf Q}^{a}_{i} - {\bf Q}^{b}_{ i}).
	\end{equation}
	Integrating Eqs.~(\ref{qQ1}), (\ref{qQ2}) and (\ref{A})-- (\ref{argmin1}), we can obtain the detailed expression of $\mathrm{F}(\cdot)$ in Eq.~(\ref{ratio}):
	\begin{equation}\label{f}
		\begin{aligned}
			\dfrac{1}{2} \| {\bf Q}^{a}_{i} - {\bf Q}^{b}_{ i}\|_{\rm 2}^{\rm 2} &= \dfrac{1}{2}\alpha_i (D^{a}_{ i})^2 - \beta_i (D^{a}_{ i}D^{b}_{ i}) + \gamma_i (D^{a}_{ i}{S})\\ &+ \dfrac{1}{2}\delta_i (D^{b}_{ i})^2 - \epsilon_i (D^{b}_{ i}{S}) + \dfrac{1}{2}\zeta_i {S}^2\\ &=\mathrm{F} \, (D^{a}_{ i},  D^{b}_{ i},{S}),
		\end{aligned}
	\end{equation}
	obviously, $ \mathrm{F} $ is a convex function. For the objective function Eq.~(\ref{ratio}), by computing the derivative of the $\rm F$ with respect to the $D^{a}_{ i}$, $D^{b}_{ i}$, ${S}$ and setting to zero, we have a $ 3N \times (2N + 1) $ linear system:
	
	\begin{equation}\label{linear system}
		\begin{pmatrix}
			\vspace{1ex}
			\alpha_1& -\beta_1 & 0& 0& \cdots   &\gamma_1\\ \vspace{1ex}
			-\beta_1& \delta_1 & 0& 0&\cdots    &-\epsilon_1\\ \vspace{1ex}
			\gamma_1& -\epsilon_1& 0& 0&\cdots & \zeta_1\\ \vspace{1ex}
			0& 0&\alpha_2& -\beta_2 &  \cdots   &\gamma_2\\ \vspace{1ex}
			0& 0&-\beta_2& \delta_2 & \cdots    &-\epsilon_2\\ \vspace{1ex}
			0& 0&\gamma_2& -\epsilon_2& \cdots & \zeta_2\\ \vspace{1ex}
			\vdots & \vdots &  & \ddots & & \vdots\\ \vspace{1ex}
			0& \cdots& \cdots&\alpha_N& -\beta_N     &\gamma_N\\ \vspace{1ex}
			0& \cdots& \cdots&-\beta_N& \delta_N    &-\epsilon_N\\ \vspace{1ex}
			0& \cdots& \cdots&\gamma_N& -\epsilon_N & \zeta_N\\ 
		\end{pmatrix}
		\begin{pmatrix}
			\vspace{1ex}
			D^{a}_{1}\\ \vspace{1ex}
			D^{b}_{1}\\ \vspace{1ex}
			D^{a}_{2}\\ \vspace{1ex}
			D^{b}_{2}\\ \vspace{1ex}
			\vdots\\ \vspace{1ex}
			\vdots\\ \vspace{1ex}
			D^{a}_{N}\\ \vspace{1ex}
			D^{b}_{N}\\ \vspace{1ex}
			S\\ 
		\end{pmatrix}
		=
		\begin{pmatrix}
			\vspace{1ex}
			0\\ \vspace{1ex}
			0\\ \vspace{1ex}
			0\\ \vspace{1ex}
			0\\ \vspace{1ex}
			0\\ \vspace{1ex}
			0\\ \vspace{1ex}
			\vdots\\ \vspace{1ex}
			0\\ \vspace{1ex}
			0\\ \vspace{1ex}
			0\\ 
		\end{pmatrix}.
	\end{equation}
	Hence, the minimum non-zero solution of Eq.~(\ref{linear system}) is the ratio between each depth and the absolute scale value.
	
	
	{\small
		\bibliographystyle{spbasic}
		\bibliography{II-ACR_ref}
	}
	
\end{document}